\documentclass[runningheads]{llncs}

\usepackage{eccv}

\usepackage{eccvabbrv}

\usepackage{graphicx}
\usepackage{booktabs}

\usepackage[accsupp]{axessibility}  

\usepackage{multirow}
\usepackage{adjustbox}
\usepackage[table]{xcolor}
\newlength{\itemwidth}

\usepackage{hyperref}

\usepackage{orcidlink}

\usepackage[capitalize]{cleveref}
\crefname{section}{Sec.}{Secs.}
\Crefname{section}{Section}{Sections}
\Crefname{table}{Table}{Tables}
\crefname{table}{Tab.}{Tabs.}

\newcommand{\Fref}[1]{Fig.~\ref{#1}}
\newcommand{\Tref}[1]{Table~\ref{#1}}

\newcommand{\capcell}[1]{%
  \parbox[t]{\itemwidth}{\centering\scriptsize #1}%
}

\begin{document}

\title{\texorpdfstring{FillGS: Filling Observation Gaps in 4D Gaussian Splatting via Viewpoint-Time Selection\\and Generative Refinement}{FillGS: Filling Observation Gaps in 4D Gaussian Splatting via Viewpoint-Time Selection and Generative Refinement}}

\titlerunning{FillGS}

\author{Takashi Otonari\orcidlink{0009-0004-1665-329X} \and
Toshihiko Yamasaki\orcidlink{0000-0002-1784-2314}}

\authorrunning{T. Otonari and T. Yamasaki}

\institute{The University of Tokyo, Japan\\
\email{\{otonari,yamasaki\}@cvm.t.u-tokyo.ac.jp}}

\maketitle

\begin{abstract}
4D Gaussian Splatting (4DGS) can render dynamic scenes photorealistically. However, with limited viewpoint coverage, some spatiotemporal regions remain sparsely observed, leading to artifacts, particularly in scenes with large motion. Existing approaches leveraging generative models rely on heuristic virtual-viewpoint selection before refining rendered views. As a result, they cannot actively explore such sparsely observed regions. To address this issue, we propose a pipeline that actively selects spatiotemporal virtual viewpoints to improve 4DGS reconstruction. Our method selects virtual viewpoints for generative enhancement based on the rendering sensitivity and motion-aware observation density of 4D Gaussians, prioritizing views that alleviate observation sparsity. In the refined images, we filter out regions that conflict with captured observations or are likely to contain generative artifacts and then fine-tune 4DGS using only the reliable regions. We evaluate our method on multi-view video benchmarks using new train/test splits designed to induce observation gaps. Results show consistent improvements over prior viewpoint selection strategies and fine-tuning methods in both qualitative and quantitative evaluations, while reducing artifacts. \footnote{Our project page is available at \href{https://otonari726.github.io/fillgs/}{https://otonari726.github.io/fillgs/}.}

\end{abstract}

\begin{figure*}[!t]
    \centering
    \setlength{\tabcolsep}{0.02cm}
    \setlength{\itemwidth}{2.4cm}
    \renewcommand{\arraystretch}{0.5}
    \begin{adjustbox}{width=0.99\linewidth}
    \hspace*{-\tabcolsep}\small\begin{tabular}{cccccc}
            \includegraphics[width=\itemwidth]{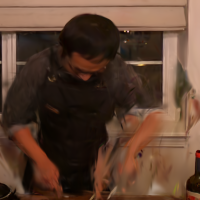} &
            \includegraphics[width=\itemwidth]{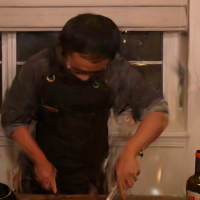} &
            \includegraphics[width=\itemwidth]{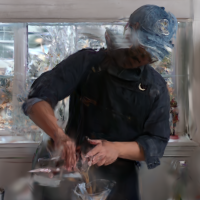} &
            \includegraphics[width=\itemwidth]{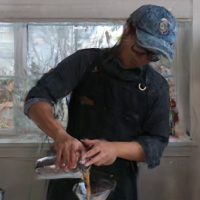} &
            \includegraphics[width=\itemwidth]{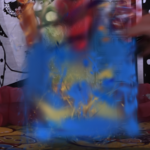} &
            \includegraphics[width=\itemwidth]{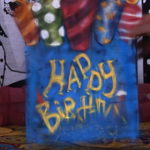} \\
            \vspace{0.2em}
            E-D3DGS~\cite{bae2024ed3dgs} &
            Ours &
            E-D3DGS~\cite{bae2024ed3dgs} &
            Ours &
            E-D3DGS~\cite{bae2024ed3dgs} &
            Ours
        \\
    \end{tabular}
    \end{adjustbox}
    \vspace{0.1cm}\text{(a) Novel views in sparse settings.}\vspace{0.1cm}
    \includegraphics[width=0.99\linewidth]{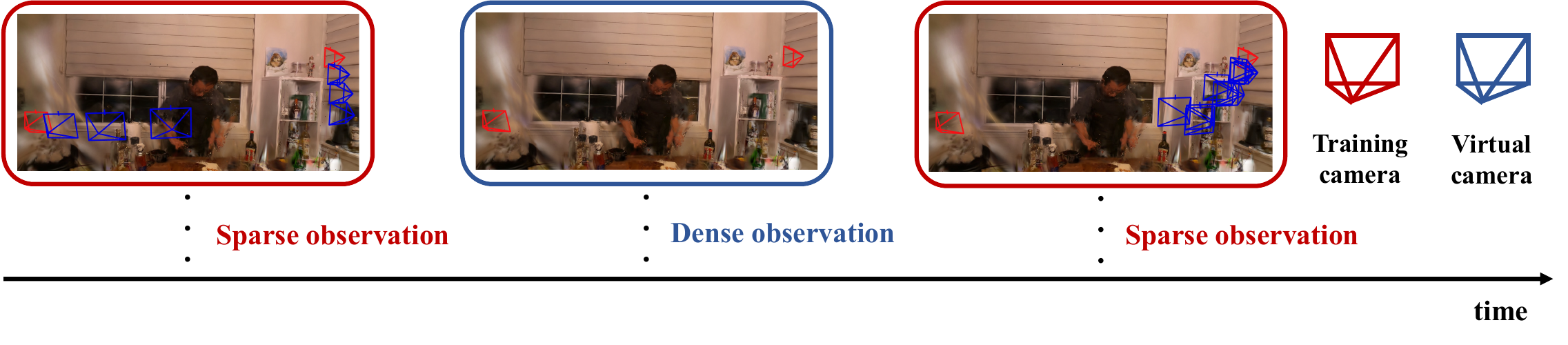}
    \vspace{0.2cm}\text{(b) Our virtual-viewpoint selection based on motion-aware observation sparsity.}
    \vspace{-0.45cm}\caption{4D Gaussian Splatting (4DGS) suffers from limited spatiotemporal observations. In this paper, we address the challenge of insufficient spatiotemporal observations by actively selecting viewpoints to identify spatiotemporal gaps and then refining the rendered images with a generative model. Our method enables high-fidelity rendering from novel spatiotemporal viewpoints.}
    \label{fig:teaser}
\end{figure*}

\section{Introduction}
\label{sec:intro}
Recent advances in neural rendering, such as Neural Radiance Fields (NeRF)~\cite{nerf}, have enabled high-fidelity novel view synthesis from multi-view images. Moreover, 3D Gaussian Splatting (3DGS)~\cite{kerbl3Dgaussians} and its dynamic extension, 4D Gaussian Splatting (4DGS)~\cite{Wu_2024_CVPR}, use explicit representations to achieve both high rendering quality and fast rendering speed, making these methods promising foundations for real-time rendering of dynamic scenes. These techniques support a wide range of applications, including augmented and virtual reality, gaming, and autonomous driving.

However, 4DGS is highly sensitive to the distribution of observed viewpoints, and thus often exhibits artifacts in sparsely observed regions, as shown in~\Fref{fig:teaser}(a). When each time step is captured by only a few cameras, fast-moving objects are seen at a given location only briefly; as a result, the corresponding 4D Gaussians are weakly constrained. In real-world captures, where viewpoint diversity is often limited, this effect is further amplified, leading to degraded reconstruction quality.

Recent studies have aimed to improve 4D novel view synthesis by leveraging generative models to refine rendered images, thereby compensating for sparse observations~\cite{4dgswild,kong2025gsgs}. However, the effectiveness of such enhanced images critically depends on the choice of virtual viewpoints used for generation; existing methods do not actively select them. Instead, they typically rely on heuristic strategies that give little consideration to the downstream effects of the generated views, such as interpolating between training viewpoints or uniformly sampling from the viewpoint space. In static scenes, selecting virtual viewpoints to maximize spatial coverage has been shown to be effective~\cite{kim2025exploregsexplorable3dscene}. However, extending this idea to dynamic scenes is fundamentally challenging because sparsity evolves over time. As motion continuously changes which regions are sparsely observed, observation sparsity is inherently spatiotemporal rather than purely spatial. Consequently, virtual viewpoints chosen solely to maximize spatial coverage fail to target sparsely observed spatiotemporal regions, especially with large motion. Accordingly, a key open problem in dynamic-scene settings is how to select spatiotemporal viewpoints for refinement by generative models.

In this paper, we propose FillGS, which actively selects spatiotemporal virtual viewpoints for the generative enhancement of 4DGS-rendered views, as shown in \Fref{fig:teaser}(b). FillGS improves rendering quality under sparse-observation settings without relying on hand-crafted heuristics, such as interpolation or uniform sampling, that fail to account for the effects of generative enhancement. We define a motion-aware observation density to identify sparsely observed regions, and we prioritize spatiotemporal regions where the 4D Gaussians are under-constrained due to sparse observations. Furthermore, we do not treat generated images as equivalent to real observations. We exclude (i) regions that are already sufficiently observed and (ii) regions likely to contain generative artifacts that are inconsistent with real observations. We then fine-tune 4DGS using only reliable pixels. As a result, FillGS enables stable improvement in sparsely observed regions without degrading already well-reconstructed regions.

To validate the proposed method, we create new training and evaluation splits from existing multi-view video datasets, thereby enabling evaluation in sparse-observation settings. Across multiple datasets under sparse-view settings, our method consistently outperforms existing viewpoint selection strategies and fine-tuning methods both qualitatively and quantitatively. 

\section{Related Work}
\label{sec:related}

\subsection{4D Novel View Synthesis}
Dynamic novel view synthesis (NVS) is considerably more challenging than static NVS because variations in space and time occur simultaneously. Among dynamic NeRF methods, many frameworks have been proposed that represent space and time as continuous functions and model scene motion via deformation fields, flow fields, or mappings to a canonical space~\cite{pumarola2020d,park2021nerfies,park2021hypernerf,li2020neural,li2023dynibar,Wang2021NeuralTF,xian2021space,Cao2023HEXPLANE}. Although these approaches achieve high-quality rendering, they incur significant optimization overhead. In addition, limited viewpoint coverage exacerbates artifacts in sparsely observed regions.

In recent years, 3D Gaussian Splatting (3DGS)~\cite{kerbl3Dgaussians} has emerged as a promising approach that achieves both high visual quality and fast rendering through an explicit point-based scene representation. Building on this line of work, 4D Gaussian Splatting (4DGS), which extends 3DGS to dynamic scenes, has advanced rapidly~\cite{4drotorgs,kratimenos2024dynmf,katsumata2024compact,lin2024gaussian,yang2023deformable3dgs,bae2024ed3dgs,lu2024gagaussian,Wu_2024_CVPR,yang2023gs4d,li2023spacetime}. Furthermore, subsequent studies have addressed challenging observation settings, such as monocular or casually captured videos~\cite{liu2025modgs,som2024,lei2024mosca,stearns2024marbles,wang2025gflow,liang2025himor,4dgswild,usplat4d2025}.

In NVS for dynamic scenes under sparse-view settings, artifacts caused by insufficient observations and inconsistencies in spatiotemporal geometric reconstruction become more pronounced. Consequently, many studies have proposed various priors and regularization techniques to mitigate these issues~\cite{kong2025gsgs,li2025gc4dgs,splatography,sparse4dgs,liu2025modgs}. In research closely related to ours, UA-4DGS~\cite{4dgswild} and GS-GS~\cite{kong2025gsgs} augment training data by using diffusion models to refine rendered images from unobserved viewpoints, compensating for sparse observations in monocular or sparse-view settings. However, the virtual viewpoints used to refine these rendered images are chosen via random sampling or interpolation between training views. As a result, these methods do not perform active virtual-viewpoint selection targeting spatiotemporal regions that are likely to improve reconstruction quality.

\subsection{View Selection for Applying Generative Priors in Novel View Synthesis}
Recent works have improved 3DGS under sparse observations by leveraging diffusion models as generative priors. In these approaches, a diffusion model refines rendered images, and the refined outputs are used as pseudo-observations to further fine-tune the reconstruction model. However, little attention has been paid to which viewpoints are effective for refining images. Existing approaches rely on heuristics such as interpolating between training views~\cite{liu20243dgs,nazarczuk2025vidar,yin2025gsfixer}, optimizing for target viewpoints~\cite{wu2025difix3d+,Wu2025GenFusion,bose2025uncertainty}, manually selecting virtual viewpoints for generation~\cite{gsfix3d}, expanding the field of view at training viewpoints~\cite{gamo2024}, and interpolating time in dynamic scenes~\cite{xiao2025vdegaussianvideodiffusionenhanced}. These designs largely ignore the viewpoint-dependent effectiveness of generated images for improving reconstruction.

In contrast, ExploreGS~\cite{kim2025exploregsexplorable3dscene} improves static reconstruction by selecting virtual viewpoints to maximize spatial coverage under sparse observations. However, dynamic scenes require spatiotemporal exploration, where observations are temporally sparse. Consequently, spatial-coverage-based selection may overlook viewpoints that are informative for motion reconstruction, thereby reducing motion estimation accuracy and degrading 4D reconstruction quality.

\subsection{Active View Selection for Data Acquisition}
To achieve high-fidelity NVS from a small number of input images, active view selection (AVS), which aims to identify informative camera views or view subsets, has been widely studied for both NeRF and 3DGS. Existing frameworks select viewpoints expected to yield the greatest improvement using criteria such as reconstruction uncertainty, information gain, scene coverage, or Fisher information~\cite{Jiang2023FisherRF,10.2312:vmv.20231222,pan2022activenerf,stochastic,densityawarenerf,NeRFDirector,popgs,wang2024avs,safadoustWarpRF,thesemagicmoments,diversitydriven}. Moreover, in robotics and online mapping, recent works have integrated active reconstruction with path planning based on NeRF or 3DGS, enabling efficient viewpoint acquisition under sensing and motion constraints~\cite{neunbv,beyonduncertainty,marza2024autonerf,Activermap,li2025activesplat}. However, many AVS methods assume that additional real observations can be obtained by capturing new images. In contrast, FillGS actively selects virtual viewpoints to refine the corresponding rendered views using a generative model over a broad spatiotemporal domain, including past time steps that are no longer physically observable.

\section{Method}
\label{sec:method}

\begin{figure*}[!t]
    \centering
    \includegraphics[width=\linewidth]{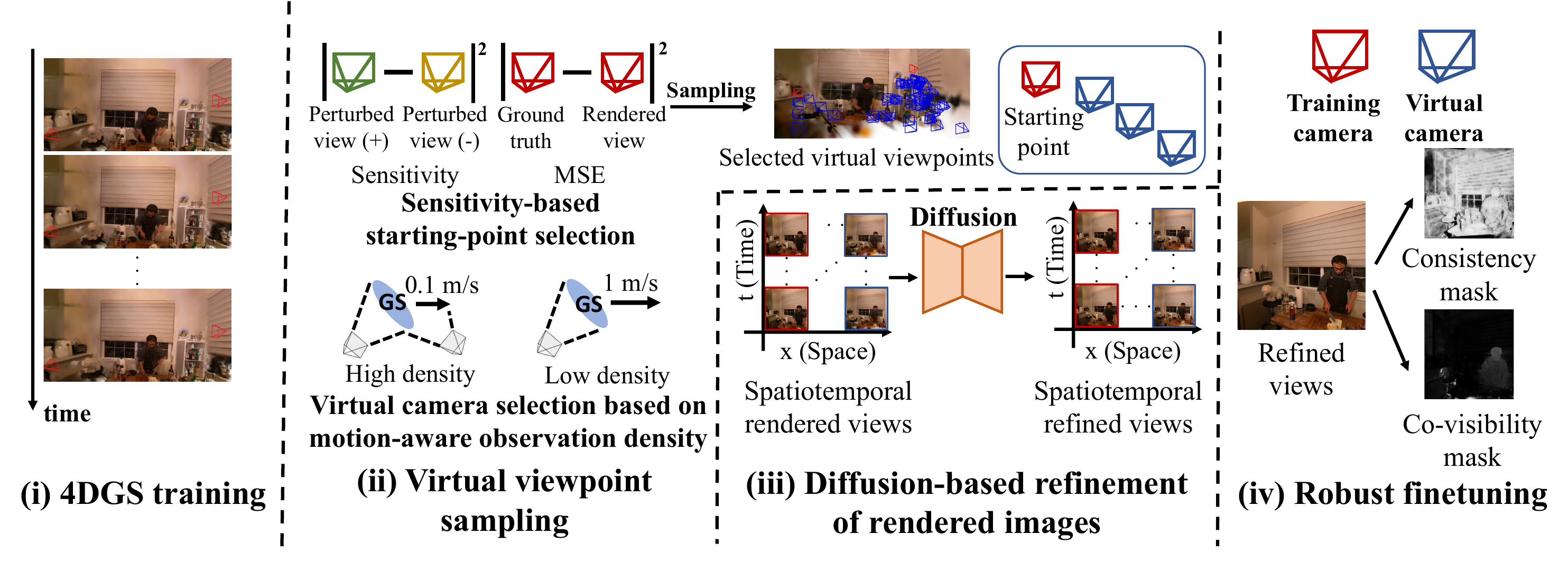}
    \vspace{-0.7cm}\caption{\textbf{Overview of our FillGS pipeline.} (i) We first train a 4DGS model. (ii) We then actively select virtual viewpoints. To do so, we first identify training views that are locally unstable due to sparse observations (Sec.~\ref{sec:sensitivity}), and then generate candidate virtual viewpoints using both local and global strategies. Each candidate is scored using a motion-aware observation-density metric, and we prioritize candidates with low observation density (Sec.~\ref{sec:uncertainty}). (iii) The rendered images from the selected virtual viewpoints are refined using a diffusion model (Sec.~\ref{sec:diffusion}). (iv) Finally, we assign weights to the refined images based on their conflicts with the training images and generation errors, and use these weights to fine-tune the 4DGS model (Sec.~\ref{sec:robust_update}).}
    \label{fig:pipeline}
\end{figure*}

\subsection{Overview}
When 4DGS is trained from a limited set of viewpoints, sparsely observed spatiotemporal regions become under-constrained, often leading to rendering artifacts. We address this by using images refined by a video diffusion model as auxiliary supervision. Our pipeline actively selects spatiotemporal virtual viewpoints that cover under-observed regions, refines the rendered images at these viewpoints, and robustly fine-tunes 4DGS while suppressing generative artifacts and conflicts with real observations. \Fref{fig:pipeline} summarizes the overall pipeline.

Uniformly sampling unobserved viewpoints does not necessarily improve the conditioning of 4DGS optimization. Instead, virtual viewpoints should be prioritized according to how well they observe regions with insufficient observation support. We therefore perform viewpoint selection in two stages. First, we select an observed spatiotemporal viewpoint where the trained 4DGS exhibits high local sensitivity, which serves as a starting point for exploration (Sec.~\ref{sec:sensitivity}). Second, we select virtual viewpoints that observe many 4D Gaussians with low motion-aware observation density (Sec.~\ref{sec:uncertainty}).

Since the goal of 4D NVS is to remain consistent with real observations, generated images are used only as auxiliary supervision for under-constrained regions. During fine-tuning, we down-weight losses from generated images in regions that are either unreliable or already well supported by observed views, using generation-error detection and motion-aware co-visibility weighting (Sec.~\ref{sec:robust_update}). This allows our method to improve sparsely observed regions without degrading well-reconstructed areas.

\subsection{Sensitivity-Based Starting-Viewpoint Selection Strategy}
\label{sec:sensitivity}
We select observed spatiotemporal viewpoints as starting points for exploration by identifying viewpoints where the trained 4DGS exhibits local instability. These viewpoints provide useful starting points for subsequent exploration.

\noindent\textbf{Why sensitivity indicates good starting points.}
We aim to select observed viewpoints whose local neighborhoods are likely to be unstable as starting points for exploration. However, because ground-truth images are unavailable for nearby unobserved views, we cannot directly evaluate reconstruction errors in these neighborhoods. We therefore approximate error-prone neighborhoods by measuring the sensitivity of the rendered output to small perturbations, using local variation of $\hat I_\theta$ as a surrogate uncertainty signal. Such weakly constrained regions often arise from insufficiently observed 4D Gaussians, where small changes in viewpoint or time can alter contributing Gaussians and visibility, causing abrupt rendering changes. Since under-constrained regions may also yield large errors even at observed views, we additionally use the squared error at observed viewpoints as an auxiliary indicator. We prioritize observed viewpoints with high error and large perturbation-induced fluctuations as starting points. We provide an analysis of this score on synthetic data in the supplementary materials.

\noindent\textbf{Score definition.}
For an observed viewpoint $v$, we compute an exploration priority score
\begin{equation}
    \mathcal{E}(v) = \mathcal{L}(v) + \mathcal{V}(v), \label{eq:sensitivity}
\end{equation}
and prioritize viewpoints with higher $\mathcal{E}(v)$ as exploration starting points. Here, $\mathcal{L}(v)$ measures the normalized reconstruction error of the observed image, and $\mathcal{V}(v)$ measures the normalized local variation of the rendered image under small perturbations. Specifically, we define
\begin{equation}
    \mathcal{L}(v) = 
    \frac{1}{\bar{\mathcal{L}}}
    \left\| 
    \hat{I}_{\theta}(v) - I(v)
    \right\|^{2},
    \label{eq:per_view_mse}
\end{equation}
where $\bar{\mathcal{L}}$ is a normalization constant chosen such that the dataset-level mean of $\mathcal{L}(v)$ is 1. This normalization ensures that $\mathcal{L}(v)$ is properly scaled and directly comparable to $\mathcal{V}(v)$. For the local variation term, we define
\begin{equation}
    \mathcal{V}(v) =
    \sum_{\delta\in\mathcal{D}}
    w_{\delta}\,
    \left\|
    \hat{I}_{\theta}(v\oplus\delta)
    -
    \hat{I}_{\theta}(v\ominus\delta)
    \right\|^{2},
    \label{eq:fd_sens}
\end{equation}
where $\mathcal{D}$ denotes a set of small perturbations (\eg, three translations, three rotations, and one temporal shift), and $v\oplus\delta$ ($v\ominus\delta$) represents the viewpoint obtained by applying perturbation $\delta$ to viewpoint $v$ in the positive (negative) direction. The weights $w_\delta$ are normalized across perturbations so that the dataset-level average contribution of each $\delta$ in the score is $1/|\mathcal{D}|$. As a result, the total over all $\delta \in \mathcal{D}$ averages to 1, matching the scale of $\mathcal{L}(v)$.

\noindent\textbf{Connection to uncertainty estimation.}
The proposed sensitivity term $\mathcal{V}(v)$ can be interpreted as a finite-difference approximation to a local Lipschitz constant of the rendering function with respect to viewpoint perturbations. Regions that are weakly constrained by observations tend to exhibit higher sensitivity to small pose or temporal changes, as minor perturbations may alter Gaussian visibilities and the blending weights. Under such conditions, the rendering function is locally unstable, and this local instability tends to correlate with epistemic uncertainty in the reconstruction. Therefore, the sensitivity score serves as a practical proxy for identifying regions with high uncertainty in the reconstruction without requiring explicit probabilistic modeling.

\subsection{Viewpoint Scoring Based on Motion-Aware Observation Density}
\label{sec:uncertainty}
In this paper, we employ two strategies to generate candidate virtual viewpoints. The first is a local strategy that searches within the neighborhood of the observed viewpoints. The second is a global strategy that samples candidate goal viewpoints directly. We then compute a motion-aware observation density based on the velocity of each 4D Gaussian and the number of observations. We assign higher scores to candidates that observe Gaussians with lower observation densities and select the highest-scoring candidate.

\noindent\textbf{Candidate view creation.}
We generate candidate virtual viewpoints to refine rendered images using (i) a local strategy that perturbs the current pose and (ii) a global strategy that samples a goal viewpoint and interpolates a trajectory from the start to that viewpoint, thereby covering regions that are difficult to reach through local exploration.

In the local approach, we generate candidate virtual viewpoints in the neighborhood of the current viewpoint by applying six translational and four rotational perturbations. In addition, we select two training viewpoints near the current viewpoint and include an interpolated viewpoint between the current viewpoint and each selected training viewpoint. In total, we obtain 12 candidate viewpoints. By incorporating interpolation with training viewpoints, the candidate set not only explores the local neighborhood but also fills gaps toward the training viewpoints, leveraging the benefits of prior interpolation-based viewpoint selection methods.

In the global approach, we create a bounding box computed from the selected starting viewpoint and its neighboring observed viewpoint at that timestamp. We then expand this bounding box by the maximum displacement reachable by the local strategy, which we calculate from the number of steps and the local search trajectory length, and uniformly sample 12 candidate viewpoints within the expanded box. For each sampled viewpoint, we compute the viewing direction by interpolating between the two neighboring observed viewpoints associated with the starting viewpoint's timestamp. By selecting candidates in regions that are difficult to reach by the local strategy, we improve global coverage. Once a global viewpoint is selected, we re-select, at the same timestamp, the training camera closest to it and construct a camera trajectory by interpolating between the selected global viewpoint and this nearest training camera. We adopt this interpolated path as the virtual-camera trajectory.

\noindent\textbf{Viewpoint scoring.}
We assign scores to the viewpoint candidates generated by the local and global strategies, and select the highest-scoring candidate as a virtual viewpoint to be used as input to a generative model. First, for each 4D Gaussian $g_i$, we construct an observation-count table $C_{g_i}(t)$ over time $t$, which indicates how many times the 4D Gaussian is observed at each time $t$. In general, a Gaussian with a large $C_{g_i}(t)$ can be regarded as having a high observation density, whereas a Gaussian with a small $C_{g_i}(t)$ can be regarded as having a low observation density. However, while the background tends to be observed consistently across all times, dynamic objects may become invisible at certain time steps or undergo substantial appearance changes, and thus at a given spatial location they are often observable only at specific times. For this reason, simple observation counts alone cannot adequately reflect the degree of motion-aware observation sparsity.

Therefore, we define a motion-aware observation-deficiency score that captures how sparsely each Gaussian is observed while accounting for its motion. Specifically, let the magnitude of the velocity of Gaussian $g_i$ at time $t$ be $s_{g_i}(t)$, and define the observation deficiency $D_{g_i}(t)$, which corresponds to the inverse of the observation density, as
\begin{equation}
D_{g_i}(t)=\frac{s_{g_i}(t)}{1+C_{g_i}(t)}.
\label{eq:motion_aware_density}
\end{equation}
According to Eq.~\eqref{eq:motion_aware_density}, a Gaussian with few observations and high speed receives a high deficiency score, indicating that it is insufficiently constrained by the captured views. In contrast, a Gaussian with many observations and low speed is assigned a low deficiency score, indicating that it is already sufficiently constrained by the captured views. For each candidate viewpoint, we render $D_{g_i}(t)$ in place of the color of each Gaussian, and take the mean pixel value of the resulting image as the candidate's score. That is, a candidate that observes more Gaussians with higher observation-deficiency scores receives a higher score, enabling the selection of highly informative viewpoints.

\noindent\textbf{Relation to information gain.}
From an active-view-selection perspective, $D_{g_i}(t)$ can be interpreted as a proxy for the information gained by observing Gaussian $g_i$ at time $t$. Gaussians exhibiting large motion while being infrequently observed tend to be less constrained, so additional observations are expected to reduce parameter uncertainty. By prioritizing viewpoints that cover such Gaussians, our scoring strategy aims to alleviate reconstruction ambiguity. 

\subsection{Video Diffusion Model}
\label{sec:diffusion}
We leverage a video diffusion model to refine images rendered from virtual viewpoints, which often exhibit artifacts when observations are sparse. The model builds on 3DGS-Enhancer~\cite{liu20243dgs} and extends it to handle dynamic scenes. Specifically, we use spatiotemporal patches as the model’s input and output. Each patch is formed by concatenating 3 consecutive frames in time and 4 views along the spatial dimension. To construct a spatiotemporal patch, we start with an observed image, append rendered images from the selected virtual viewpoints, and extend the sequence to adjacent time steps in both temporal directions. As additional conditioning inputs, we provide the camera parameters and timestamps corresponding to each observed and rendered frame. Conditioned on these inputs, the model predicts a refined image sequence with completed content and improved spatiotemporal consistency.

Training follows the procedure of 3DGS-Enhancer. We initialize the model from weights pretrained on the DL3DV dataset~\cite{dl3dv} and then further fine-tune it on a dataset of dynamic scenes. Details of the dataset and the diffusion training procedure are provided in the supplementary materials.

\subsection{Robust 4D Model Fine-tuning Using Generated Views}
\label{sec:robust_update}
We propose a robust fine-tuning strategy that preserves observation consistency in 4DGS. Specifically, we modulate the loss using (i) matching-based consistency masks and (ii) motion-aware co-visibility masks, so that fine-tuning concentrates on non-conflicting, under-constrained regions.

\noindent\textbf{Consistency mask driven by matching confidence.}
To assess the reliability of the generated images, we use RoMa~v2~\cite{edstedt2025romav2} to compute geometric consistency between the observed and generated images, producing a per-pixel confidence map. For each generated image, we match it to its two adjacent observed views and take the maximum of the two confidence scores for each pixel. This assigns higher weight to regions that are observed from at least one neighboring view. This confidence map can be interpreted as the likelihood that a generated pixel is consistent with the observed views. We fine-tune the 4DGS model using the consistency mask to suppress the influence of incorrectly generated regions.

\noindent\textbf{Co-visibility mask with motion-aware observation-density weighting.}
When fine-tuning a model using generated images, it is important not to overly modify regions that are already supported by existing observations. Existing approaches for incorporating generated images into dynamic scenes either render per-view uncertainty maps from static per-Gaussian uncertainty values~\cite{4dgswild} or apply the same photometric loss weights as those used for real images~\cite{kong2025gsgs}. These weighting strategies do not explicitly account for dynamic components in the scene.

To address this issue, we leverage the motion-aware observation density of 4D Gaussians to adaptively assign weights to generated images. Specifically, we assign small weights to regions with high observation density that are well-constrained by existing observations, thereby suppressing updates, while assigning large weights to sparsely observed regions, thereby strengthening the constraint imposed by generated images. These weights are obtained by normalizing the observation deficiency score $D_{g_i}(t)$, which is also used for virtual-viewpoint selection. Consequently, we achieve stable model fine-tuning that exploits the benefits of generated images without degrading observation-consistent reconstructions.

\begin{figure*}[t!]
    \centering
    \begin{minipage}[t]{0.49\linewidth}
        \begin{minipage}[t]{0.49\linewidth}
            \centering
            \includegraphics[width=\linewidth]{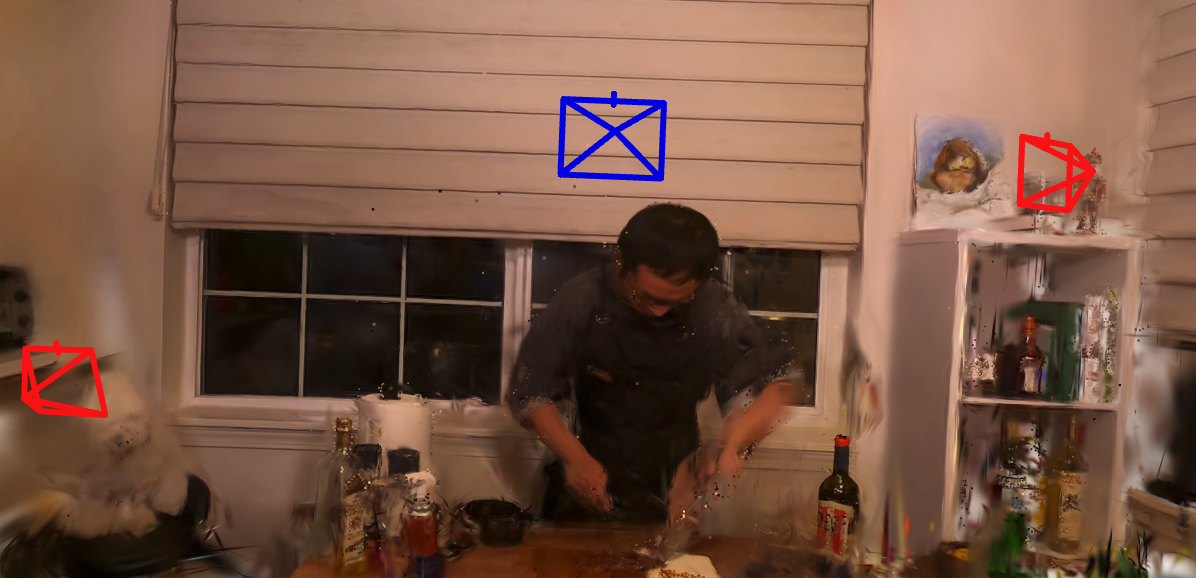}
            {\scriptsize\texttt{interpolation}}\vspace{1mm}
        \end{minipage}\hfill
        \begin{minipage}[t]{0.49\linewidth}
            \centering
            \includegraphics[width=\linewidth]{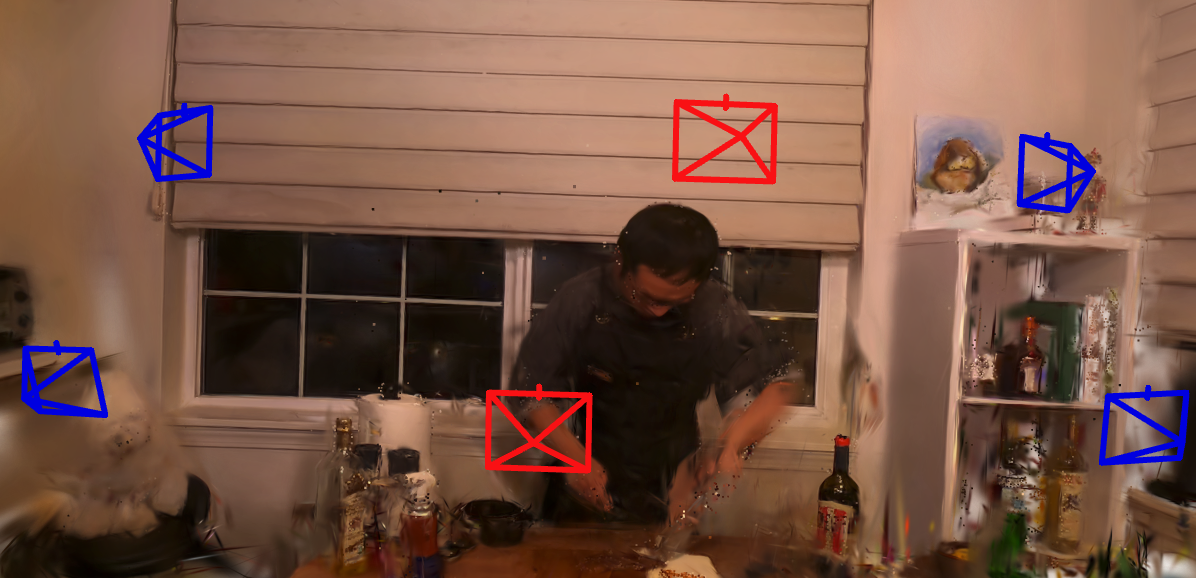}
            {\scriptsize\texttt{extrapolation}}\vspace{1mm}
        \end{minipage}\hfill
        \centering{\small\textbf{Neural 3D Video~\cite{dynerf}}}\vspace{1mm}
    \end{minipage}
    \begin{minipage}[t]{0.49\linewidth}
        \begin{minipage}[t]{0.49\linewidth}
            \centering
            \includegraphics[width=\linewidth]{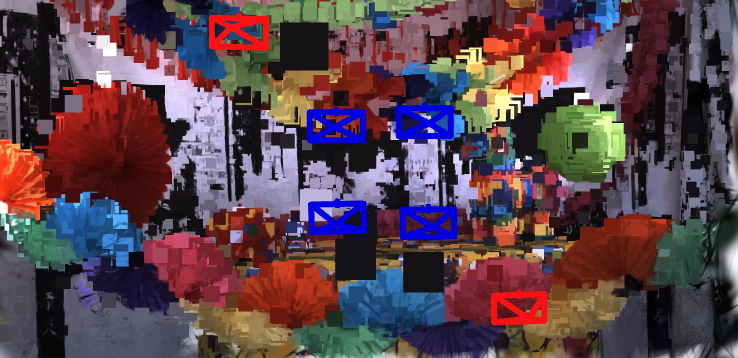}
            {\scriptsize\texttt{interpolation}}\vspace{1mm}
        \end{minipage}\hfill
        \begin{minipage}[t]{0.49\linewidth}
            \centering
            \includegraphics[width=\linewidth]{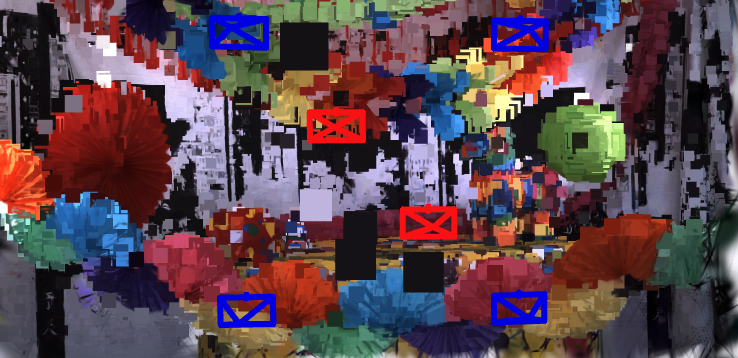}
            {\scriptsize\texttt{extrapolation}}\vspace{1mm}
        \end{minipage}
        \centering{\small\textbf{Technicolor~\cite{technicolor}}}\vspace{1mm}
    \end{minipage}
    \vspace{-0.4cm}\caption{\textbf{Our sparse-view camera split.} We train on two cameras and evaluate on spatially separated views under both interpolation and extrapolation splits. Red cameras indicate training cameras, and blue cameras indicate testing cameras.}
    \label{fig:camera_split}
\end{figure*}

\section{Experiments}
\subsection{Experimental Settings}

\noindent\textbf{Implementation details.}
We adopt E-D3DGS~\cite{bae2024ed3dgs} as our baseline, initializing it according to the original training configuration. The model is then fine-tuned for 30,000 iterations; at 6,000-iteration intervals, we generate 20 virtual trajectories and incorporate them into the fine-tuning process. All experiments are conducted using a single NVIDIA A100 GPU with either 40 GB or 80 GB of memory. Additional details are provided in the supplementary materials.

\noindent\textbf{Evaluation metrics.}
We evaluate rendering quality using standard image-level metrics, including PSNR, SSIM~\cite{ssim}, and LPIPS~\cite{lpips}. To better capture perceptual and semantic consistency, we additionally report (i) FID~\cite{fid}, which measures the distance between the feature distributions of rendered and ground-truth images, and (ii) the cosine similarity between DINOv2~\cite{dinov2} features extracted from rendered and ground-truth images.

\noindent\textbf{Baselines.}
To fairly evaluate the proposed pipeline for virtual-viewpoint selection and model fine-tuning, we control for all factors other than the viewpoint selection strategy, fine-tuning method, and the use of generated images. Concretely, we adopt E-D3DGS~\cite{bae2024ed3dgs} as a common backbone for all methods. When generated images are introduced during training, we fix both the generative model and the number of generated images. For the baseline methods, the weighting scheme for incorporating generated images follows ExploreGS~\cite{kim2025exploregsexplorable3dscene} and is computed using G-IoU and LPIPS. Under this setup, we compare the following viewpoint selection strategies proposed in prior work: (i) \textbf{E-D3DGS}~\cite{bae2024ed3dgs}, which trains without any generated images and uses only the original training images; (ii) \textbf{Interpolation-based selection}~\cite{kong2025gsgs}, which restricts virtual viewpoints to interpolations of training views; (iii) \textbf{FisherRF-based selection}~\cite{Jiang2023FisherRF} using Fisher information scores; (iv) \textbf{Coverage-based (3D) selection}~\cite{kim2025exploregsexplorable3dscene}, which follows ExploreGS for virtual-viewpoint selection by maintaining a single coverage map shared across all time steps and prioritizing views that observe more 4D Gaussians from novel directions; and (v) \textbf{Coverage-based (4D) selection}~\cite{kim2025exploregsexplorable3dscene}, which extends ExploreGS viewpoint selection to 4D by maintaining a separate coverage map for each time step, thereby prioritizing views that observe more 4D Gaussians from novel directions without considering coverage at other time steps and maximizing per-time-step spatial coverage.

\noindent\textbf{Datasets.}
Most dynamic novel view synthesis datasets are captured with synchronized multi-camera rigs, and benchmarks typically sample test views between densely spaced training cameras, which mainly evaluates interpolation. This is inadequate for sparse-view robustness, where test viewpoints lie far from the training trajectory. We therefore introduce new benchmark splits from 6 scenes in Neural 3D Video~\cite{dynerf} and 5 scenes in Technicolor~\cite{technicolor}. We train on only two cameras and evaluate on two test sets: interpolation views between the training cameras and extrapolation views outside their trajectory (\Fref{fig:camera_split}). This setting better measures generalization to sparsely observed regions.

\begin{table*}[t]
    \centering
    \begin{adjustbox}{width=1.0\linewidth}
    \small\begin{tabular}{c|rrrrr|rrrrr}
    \renewcommand{\arraystretch}{1.05}
    & \multicolumn{5}{c|}{Interpolation} & \multicolumn{5}{c}{Extrapolation} \\
    & PSNR$\uparrow$ & SSIM$\uparrow$ & LPIPS$\downarrow$ & FID$\downarrow$ & DINOv2$\uparrow$ & PSNR$\uparrow$ & SSIM$\uparrow$ & LPIPS$\downarrow$ & FID$\downarrow$ & DINOv2$\uparrow$ \\\hline
    E-D3DGS~\cite{bae2024ed3dgs} & 20.42 & 0.775 & 0.293 & 129.33 & 0.871 & 14.68 & 0.723 & 0.363 & 160.01 & 0.832 \\
    Interpolation-based selection~\cite{kong2025gsgs} & 20.75 & 0.777 & 0.293 & 120.08 & 0.896 & 14.75 & 0.716 & 0.370 & 157.52 & 0.826 \\
    FisherRF-based selection~\cite{Jiang2023FisherRF} & 21.01 & 0.774 & 0.303 & 135.36 & 0.853 & 15.59 & 0.716 & 0.373 & 180.27 & 0.800 \\
    Coverage-based (3D) selection~\cite{kim2025exploregsexplorable3dscene} & 19.91 & 0.754 & 0.317 & 148.68 & 0.878 & 15.30 & 0.700 & 0.388 & 175.53 & 0.809 \\
    Coverage-based (4D) selection~\cite{kim2025exploregsexplorable3dscene} & 19.09 & 0.737 & 0.325 & 140.19 & 0.866 & 15.51 & 0.681 & 0.407 & 182.99 & 0.792 \\
    Ours & \textbf{21.73} & \textbf{0.792} & \textbf{0.256} & \textbf{83.81} & \textbf{0.920} & \textbf{16.24} & \textbf{0.740} & \textbf{0.312} & \textbf{137.32} & \textbf{0.855} \\
    \end{tabular}
    \end{adjustbox}
    \caption{\textbf{Quantitative comparison on Neural 3D Video Dataset~\cite{dynerf}.}}
    \label{table:experiment_dynerf}
\end{table*}

\begin{table*}[t!]
    \centering
    \begin{adjustbox}{width=1.0\linewidth}
    \small\begin{tabular}{c|rrrrr|rrrrr}
    \renewcommand{\arraystretch}{1.05}
    & \multicolumn{5}{c|}{Interpolation} & \multicolumn{5}{c}{Extrapolation} \\
    & PSNR$\uparrow$ & SSIM$\uparrow$ & LPIPS$\downarrow$ & FID$\downarrow$ & DINOv2$\uparrow$ & PSNR$\uparrow$ & SSIM$\uparrow$ & LPIPS$\downarrow$ & FID$\downarrow$ & DINOv2$\uparrow$ \\\hline
    E-D3DGS~\cite{bae2024ed3dgs} & 19.63 & 0.606 & 0.378 & 137.86 & 0.736 & 16.10 & 0.690 & 0.341 & 131.48 & 0.785 \\
    Interpolation-based selection~\cite{kong2025gsgs} & 19.90 & 0.612 & 0.376 & 115.24 & 0.768 & 16.88 & 0.701 & 0.338 & 123.77 & 0.786 \\
    FisherRF-based selection~\cite{Jiang2023FisherRF} & 20.08 & 0.614 & 0.421 & 152.56 & 0.686 & 18.32 & 0.646 & 0.444 & 163.51 & 0.636 \\
    Coverage-based (3D) selection~\cite{kim2025exploregsexplorable3dscene} & 20.12 & 0.622 & 0.370 & 106.63 & 0.770 & 17.61 & 0.710 & 0.332 & 110.54 & 0.805 \\
    Coverage-based (4D) selection~\cite{kim2025exploregsexplorable3dscene} & 20.03 & 0.619 & 0.373 & 110.07 & 0.769 & 17.68 & 0.717 & 0.329 & 106.47 & 0.811 \\
    Ours & \textbf{22.55} & \textbf{0.748} & \textbf{0.250} & \textbf{73.44} & \textbf{0.866} & \textbf{19.88} & \textbf{0.720} & \textbf{0.315} & \textbf{106.45} & \textbf{0.821} \\
    \end{tabular}
    \end{adjustbox}
    \caption{\textbf{Quantitative comparison on Technicolor Dataset~\cite{technicolor}.}}
    \label{table:experiment_technicolor}
\end{table*}

\begin{figure*}[t]
    \centering
    \setlength{\tabcolsep}{0.02cm}
    \setlength{\itemwidth}{2.4cm}
    \renewcommand{\arraystretch}{0.5}
    \begin{adjustbox}{width=1.0\linewidth}
    \hspace*{-\tabcolsep}\small\begin{tabular}{cccccccc}
            \includegraphics[width=1.33333333333\itemwidth]{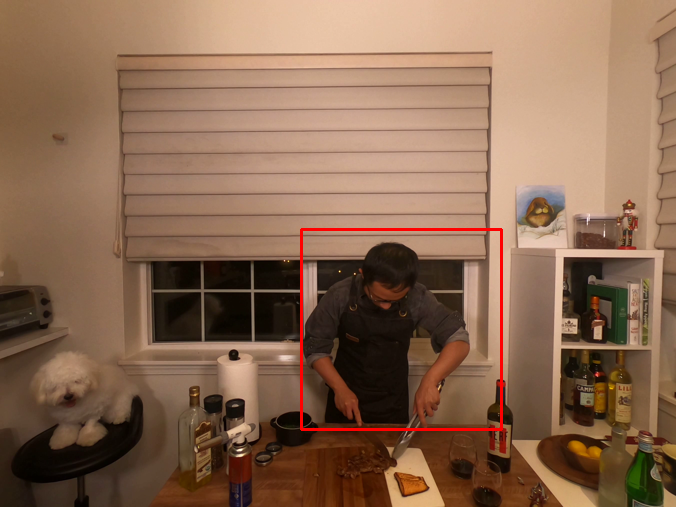} &
            \fboxsep=0pt\fcolorbox{red}{white}{\includegraphics[width=\itemwidth]{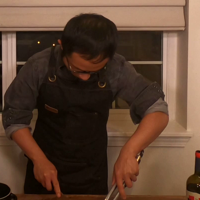}} &
            \includegraphics[width=\itemwidth]{figures/4/res/1/1_0000.png} &
            \includegraphics[width=\itemwidth]{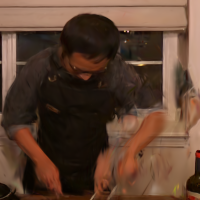} &
            \includegraphics[width=\itemwidth]{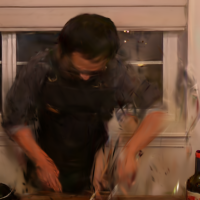} &
            \includegraphics[width=\itemwidth]{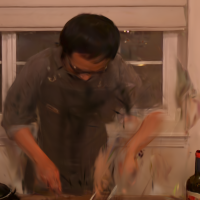} &
            \includegraphics[width=\itemwidth]{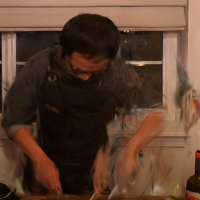} &
            \includegraphics[width=\itemwidth]{figures/4/res/1/5_0000.png} \\
            \includegraphics[width=1.33333333333\itemwidth]{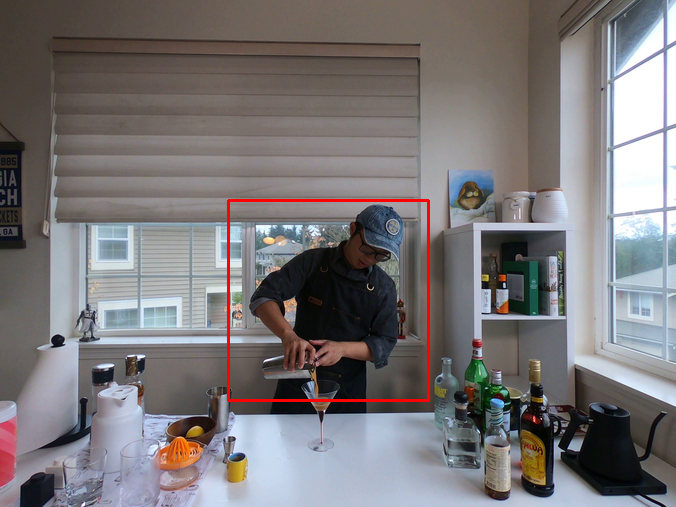} &
            \fboxsep=0pt\fcolorbox{red}{white}{\includegraphics[width=\itemwidth]{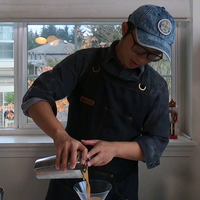}} &
            \includegraphics[width=\itemwidth]{figures/4/res/2/1_0050.png} &
            \includegraphics[width=\itemwidth]{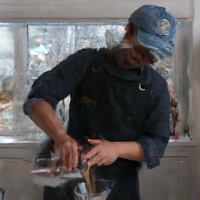} &
            \includegraphics[width=\itemwidth]{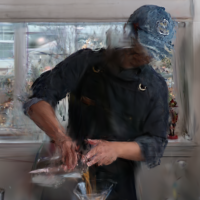} &
            \includegraphics[width=\itemwidth]{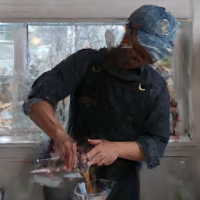} &
            \includegraphics[width=\itemwidth]{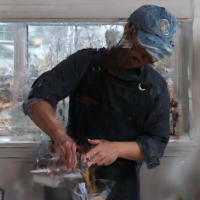} &
            \includegraphics[width=\itemwidth]{figures/4/res/2/5_0050.png} \\
            \includegraphics[width=1.33333333333\itemwidth]{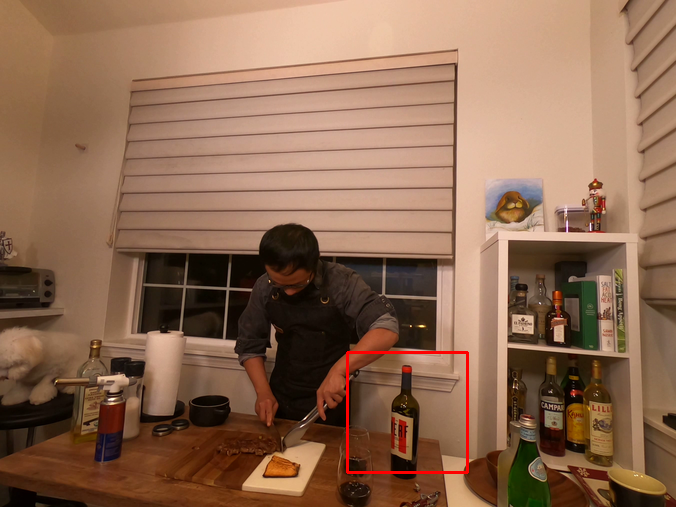} &
            \fboxsep=0pt\fcolorbox{red}{white}{\includegraphics[width=\itemwidth]{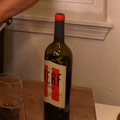}} &
            \includegraphics[width=\itemwidth]{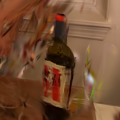} &
            \includegraphics[width=\itemwidth]{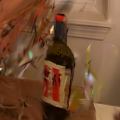} &
            \includegraphics[width=\itemwidth]{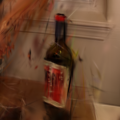} &
            \includegraphics[width=\itemwidth]{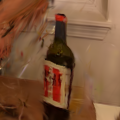} &
            \includegraphics[width=\itemwidth]{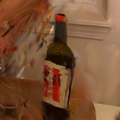} &
            \includegraphics[width=\itemwidth]{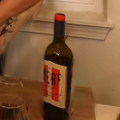} \\
            \multicolumn{2}{c}{Ground-Truth image} &
            \capcell{E-D3DGS~\cite{bae2024ed3dgs}} &
            \capcell{Interpolation-based\\selection~\cite{kong2025gsgs}} &
            \capcell{FisherRF-based\\selection~\cite{Jiang2023FisherRF}} &
            \capcell{Coverage-based\\(3D) selection~\cite{kim2025exploregsexplorable3dscene}} &
            \capcell{Coverage-based\\(4D) selection~\cite{kim2025exploregsexplorable3dscene}} &
            \capcell{Ours} \\
        \\
    \end{tabular}\vspace{-1.5em}
    \end{adjustbox}
  \vspace{-0.4cm}\caption{\textbf{Qualitative comparison on Neural 3D Video Dataset~\cite{dynerf}.}}
  \label{fig:exp_dynerf}
\end{figure*}

\begin{figure*}[t]
    \centering
    \setlength{\tabcolsep}{0.02cm}
    \setlength{\itemwidth}{2.4cm}
    \renewcommand{\arraystretch}{0.5}
    \begin{adjustbox}{width=1.0\linewidth}
    \hspace*{-\tabcolsep}\small\begin{tabular}{cccccccc}
            \includegraphics[width=1.88235294118\itemwidth]{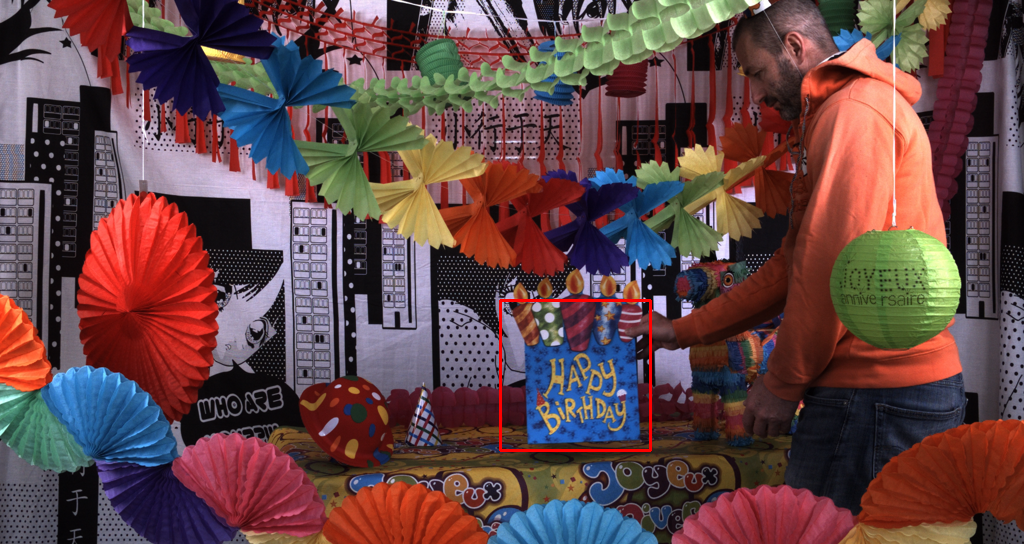} &
            \fboxsep=0pt\fcolorbox{red}{white}{\includegraphics[width=\itemwidth]{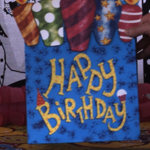}} &
            \includegraphics[width=\itemwidth]{figures/4/res/13/1_0140.png} &
            \includegraphics[width=\itemwidth]{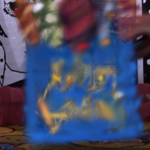} &
            \includegraphics[width=\itemwidth]{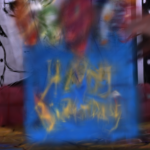} &
            \includegraphics[width=\itemwidth]{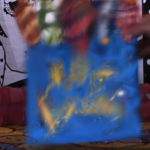} &
            \includegraphics[width=\itemwidth]{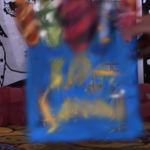} &
            \includegraphics[width=\itemwidth]{figures/4/res/13/5_0140.png} \\
            \includegraphics[width=1.88235294118\itemwidth]{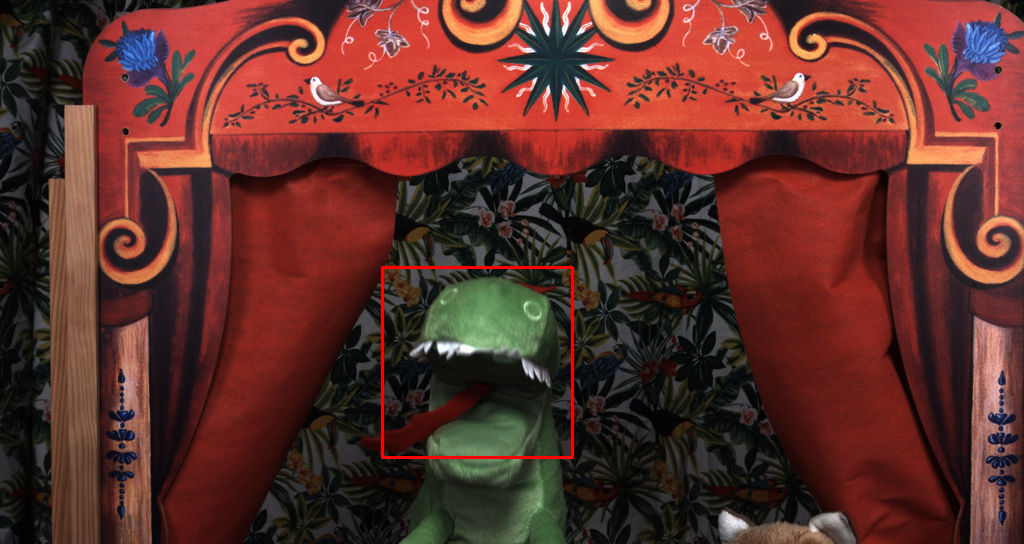} &
            \fboxsep=0pt\fcolorbox{red}{white}{\includegraphics[width=\itemwidth]{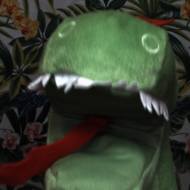}} &
            \includegraphics[width=\itemwidth]{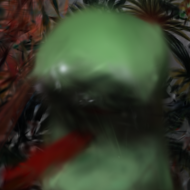} &
            \includegraphics[width=\itemwidth]{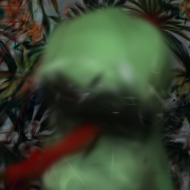} &
            \includegraphics[width=\itemwidth]{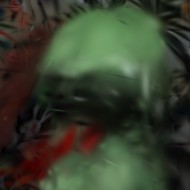} &
            \includegraphics[width=\itemwidth]{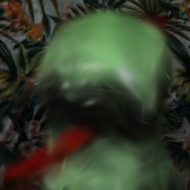} &
            \includegraphics[width=\itemwidth]{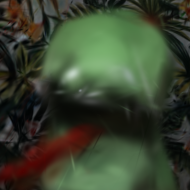} &
            \includegraphics[width=\itemwidth]{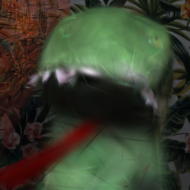} \\
            \includegraphics[width=1.88235294118\itemwidth]{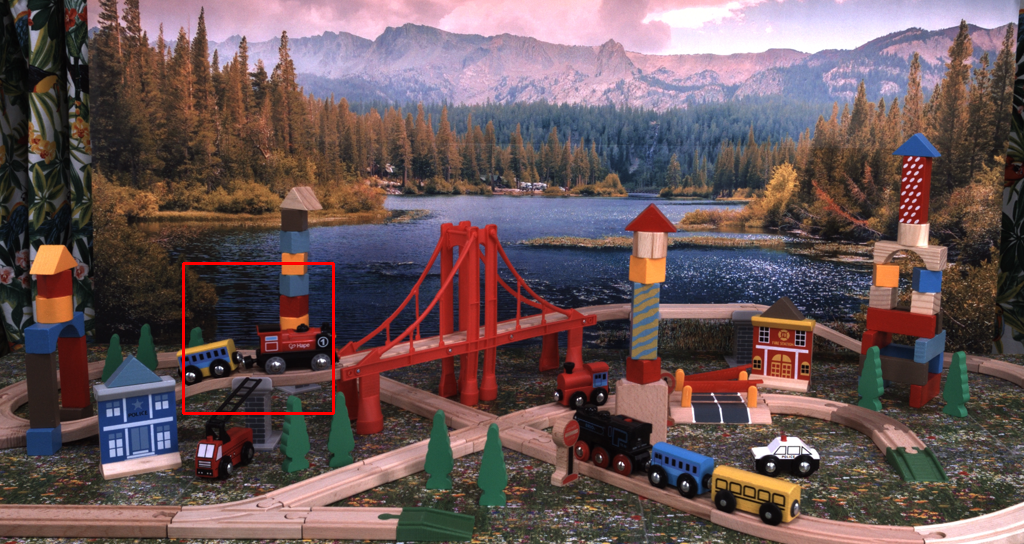} &
            \fboxsep=0pt\fcolorbox{red}{white}{\includegraphics[width=\itemwidth]{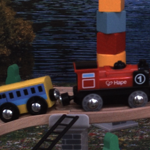}} &
            \includegraphics[width=\itemwidth]{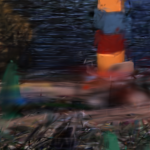} &
            \includegraphics[width=\itemwidth]{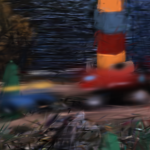} &
            \includegraphics[width=\itemwidth]{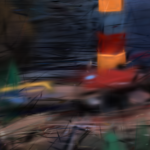} &
            \includegraphics[width=\itemwidth]{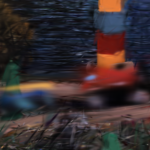} &
            \includegraphics[width=\itemwidth]{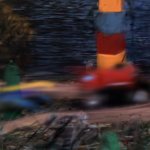} &
            \includegraphics[width=\itemwidth]{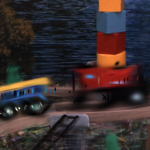} \\
            \multicolumn{2}{c}{Ground-Truth image} &
            \capcell{E-D3DGS~\cite{bae2024ed3dgs}} &
            \capcell{Interpolation-based\\selection~\cite{kong2025gsgs}} &
            \capcell{FisherRF-based\\selection~\cite{Jiang2023FisherRF}} &
            \capcell{Coverage-based\\(3D) selection~\cite{kim2025exploregsexplorable3dscene}} &
            \capcell{Coverage-based\\(4D) selection~\cite{kim2025exploregsexplorable3dscene}} &
            \capcell{Ours} \\
        \\
    \end{tabular}\vspace{-1.5em}
    \end{adjustbox}
  \vspace{-0.4cm}\caption{\textbf{Qualitative comparison on Technicolor Dataset~\cite{technicolor}.}}
  \label{fig:exp_technicolor}
\end{figure*}

\subsection{Results}
\noindent\textbf{Quantitative comparison.}
Quantitative results on the Neural 3D Video~\cite{dynerf} and Technicolor~\cite{technicolor} datasets are reported in \Tref{table:experiment_dynerf} and \Tref{table:experiment_technicolor}, respectively. Our method improves performance across all evaluation metrics and remains robust under both interpolation and extrapolation. While prior methods often focus on either interpolation between training views or extrapolation beyond them to expand viewpoint coverage, our approach performs well in both settings. Additional experiments using other datasets, generative models, and backbones are provided in the supplementary materials.

\begin{table*}[t]
    \centering
    \begin{adjustbox}{width=1.0\linewidth}
    \small\begin{tabular}{l|rrrrr|rrrrr}
    \renewcommand{\arraystretch}{1.05}
    & \multicolumn{5}{c|}{Interpolation} & \multicolumn{5}{c}{Extrapolation} \\
    & PSNR$\uparrow$ & SSIM$\uparrow$ & LPIPS$\downarrow$ & FID$\downarrow$ & DINOv2$\uparrow$ & PSNR$\uparrow$ & SSIM$\uparrow$ & LPIPS$\downarrow$ & FID$\downarrow$ & DINOv2$\uparrow$ \\\hline
    Ours & \textbf{21.73} & \textbf{0.792} & \textbf{0.256} & \textbf{83.81} & 0.920 & \textbf{16.24} & \textbf{0.740} & \textbf{0.312} & \textbf{137.32} & \textbf{0.855} \\
    \multicolumn{1}{l|}{\textit{Our starting-viewpoint selection}} & & & & & & \\
    w/o starting-viewpoint selection & 21.06 & 0.773 & 0.286 & 126.78 & 0.880 & 15.75 & 0.705 & 0.393 & 144.98 & 0.822 \\
    w/o sensitivity-based selection & 21.24 & 0.763 & 0.279 & 100.77 & \textbf{0.922} & 16.01 & 0.714 & 0.379 & 146.25 & 0.826 \\
    w/o error-based selection & 21.52 & 0.766 & 0.275 & 100.14 & 0.919 & 16.22 & 0.711 & 0.377 & 142.81 & 0.830 \\
    \multicolumn{1}{l|}{\textit{Our candidate-viewpoint selection}} & & & & & \\
    w/o candidate-viewpoint selection & 21.13 & 0.778 & 0.282 & 113.72 & 0.902 & 15.07 & 0.650 & 0.391 & 147.03 & 0.825 \\
    w/o observation-count-based selection & 20.55 & 0.768 & 0.294 & 135.20 & 0.874 & 15.66 & 0.703 & 0.346 & 150.63 & 0.833 \\
    w/o speed-based selection & 21.06 & 0.761 & 0.283 & 106.35 & 0.914 & 15.79 & 0.703 & 0.350 & 146.81 & 0.823 \\
    w/o global candidate view creation & 21.39 & 0.778 & 0.283 & 116.88 & 0.894 & 15.85 & 0.701 & 0.353 & 150.18 & 0.830 \\
    \multicolumn{1}{l|}{\textit{Other selection strategy}} & & & & & \\
    Random selection~\cite{4dgswild} & 20.59 & 0.762 & 0.295 & 130.43 & 0.869 & 15.07 & 0.701 & 0.384 & 172.37 & 0.811 \\
    Negative selection & 20.31 & 0.765 & 0.294 & 129.30 & 0.866 & 15.06 & 0.700 & 0.379 & 175.95 & 0.808 \\
    Interpolation-based selection~\cite{kong2025gsgs} & 21.15 & 0.776 & 0.283 & 120.52 & 0.888 & 15.30 & 0.703 & 0.386 & 173.37 & 0.817 \\
    FisherRF-based selection~\cite{Jiang2023FisherRF} & 20.27 & 0.761 & 0.302 & 139.41 & 0.855 & 15.21 & 0.709 & 0.363 & 157.92 & 0.823 \\
    Coverage-based (3D) selection~\cite{kim2025exploregsexplorable3dscene} & 20.94 & 0.770 & 0.290 & 133.28 & 0.870 & 15.51 & 0.705 & 0.384 & 171.75 & 0.820 \\
    Coverage-based (4D) selection~\cite{kim2025exploregsexplorable3dscene} & 20.62 & 0.771 & 0.280 & 131.06 & 0.895 & 15.69 & 0.700 & 0.388 & 167.36 & 0.819 \\
    \end{tabular}
    \end{adjustbox}
    \caption{\textbf{An ablation study of our viewpoint selection on Neural 3D Video Dataset~\cite{dynerf}.}}
    \label{table:ablation_selection}
\end{table*}

\begin{table*}[t]
    \centering
    \begin{adjustbox}{width=1.0\linewidth}
    \small\begin{tabular}{l|rrrrr|rrrrr}
    \renewcommand{\arraystretch}{1.05}
    & \multicolumn{5}{c|}{Interpolation} & \multicolumn{5}{c}{Extrapolation} \\
    & PSNR$\uparrow$ & SSIM$\uparrow$ & LPIPS$\downarrow$ & FID$\downarrow$ & DINOv2$\uparrow$ & PSNR$\uparrow$ & SSIM$\uparrow$ & LPIPS$\downarrow$ & FID$\downarrow$ & DINOv2$\uparrow$ \\\hline
    Ours & 21.73 & \textbf{0.792} & \textbf{0.256} & \textbf{83.81} & \textbf{0.920} & \textbf{16.24} & \textbf{0.740} & \textbf{0.312} & \textbf{137.32} & \textbf{0.855} \\
    \multicolumn{1}{l|}{\textit{Our fine-tuning method}} & & & & & & \\
    w/o weight & 20.05 & 0.716 & 0.334 & 128.57 & 0.891 & 14.70 & 0.639 & 0.409 & 167.40 & 0.829 \\
    w/o consistency mask & 21.26 & 0.778 & 0.283 & 107.96 & 0.897 & 15.81 & 0.705 & 0.355 & 149.50 & 0.833 \\
    w/o co-visibility mask & \textbf{21.96} & 0.780 & 0.270 & 98.43 & 0.912 & 15.93 & 0.700 & 0.359 & 153.95 & 0.839 \\
    \multicolumn{1}{l|}{\textit{Other weighting method}} & & & & & \\
    3DGS-Enhancer~\cite{liu20243dgs} & 21.41 & 0.763 & 0.281 & 100.11 & 0.910 & 15.48 & 0.694 & 0.374 & 156.29 & 0.811 \\
    ExploreGS~\cite{kim2025exploregsexplorable3dscene} & 20.97 & 0.770 & 0.287 & 109.74 & 0.895 & 15.14 & 0.697 & 0.353 & 151.38 & 0.820 \\
    UA-4DGS~\cite{4dgswild} & 19.96 & 0.745 & 0.322 & 135.11 & 0.847 & 14.69 & 0.689 & 0.419 & 201.89 & 0.772 \\
    \end{tabular}
    \end{adjustbox}
    \caption{\textbf{An ablation study of our fine-tuning methods on Neural 3D Video Dataset~\cite{dynerf}.}}
    \label{table:ablation_weight}
\end{table*}

\noindent\textbf{Qualitative comparison.}
\Fref{fig:exp_dynerf} and \Fref{fig:exp_technicolor} show qualitative comparisons on the Neural 3D Video~\cite{dynerf} and Technicolor~\cite{technicolor} datasets. In all rows of both figures, the proposed method effectively restores fine details in dynamic regions, such as thin structures and motion boundaries (\Fref{fig:exp_dynerf} top and second rows; \Fref{fig:exp_technicolor} top and second rows), while simultaneously mitigating background artifacts (\Fref{fig:exp_dynerf} bottom row; \Fref{fig:exp_technicolor} bottom row). In contrast, prior approaches often exhibit oversmoothing in fast-moving areas and residual artifacts in static background regions. These results indicate that our model enables more accurate reconstruction of geometry and appearance under challenging motion and artifact-prone conditions. Overall, the proposed method consistently achieves higher visual fidelity than existing methods. Additional visualizations of the failure cases are provided in the supplementary materials.

\subsection{Ablation Study}
\noindent\textbf{Viewpoint selection.}
To assess the effectiveness of our virtual-viewpoint selection, we compare it with several baseline strategies. The results on the Neural 3D Video dataset~\cite{dynerf} are shown in \Tref{table:ablation_selection}. Random selection chooses virtual viewpoints uniformly at random as in UA-4DGS~\cite{4dgswild}. Negative selection inverts the score used in our method.  The interpolation-based selection samples virtual viewpoints between training views. FisherRF-based selection, following FisherRF~\cite{Jiang2023FisherRF}, selects viewpoints to maximize the Fisher information score. The coverage-based selection, following ExploreGS~\cite{kim2025exploregsexplorable3dscene}, selects viewpoints to maximize spatial coverage. Our method consistently outperforms these baselines. Moreover, the performance drop with negative selection suggests that our score is a meaningful indicator. Restricting virtual-viewpoint candidate creation to local strategies limits global exploration, reduces viewpoint diversity, and consequently degrades accuracy. In addition, both starting-viewpoint and candidate-viewpoint selection are important. Removing either component reduces the likelihood of sampling informative viewpoints, leading to lower accuracy. Furthermore, the error and sensitivity scores used for starting-viewpoint selection, as well as the observation count and speed used for candidate-viewpoint selection, are all important; removing any of them degrades performance. Additional experiments on the cost–accuracy trade-off of viewpoint selection are provided in the supplementary materials.

In \Fref{fig:vis_sampling}, we visualize the spatial distribution of virtual viewpoints generated by each viewpoint selection strategy by collapsing the temporal dimension. With interpolation-based selection between the training viewpoints, the sampled virtual viewpoints cluster in limited regions. In contrast, the coverage-based selection is biased toward extrapolation, rather than filling the gaps between the training viewpoints. By contrast, our proposed method samples virtual viewpoints that effectively cover the gaps between training viewpoints while avoiding excessive extrapolation. In the extrapolation setting, the global strategy samples extrapolative viewpoints beyond the locally reachable region, mainly improving extrapolation performance, while local candidates stabilize nearby geometry. Further analysis of starting-viewpoint selection using synthetic data, including the temporal axis, is provided in the supplementary materials.

\noindent\textbf{Fine-tuning.}
To verify the effectiveness of our fine-tuning strategy, we compare it with existing fine-tuning approaches, as shown in \Tref{table:ablation_weight}. 3DGS-Enhancer~\cite{liu20243dgs} assigns weights based on the distance from the training viewpoints and Gaussian scales. ExploreGS~\cite{kim2025exploregsexplorable3dscene} uses LPIPS~\cite{lpips} between the rendered and refined images, as well as G-IoU derived from Gaussian co-visibility. UA-4DGS~\cite{4dgswild} weights samples by uncertainty computed from the rendering contribution weights on the training images. Our method consistently outperforms existing fine-tuning approaches across all metrics. In addition, while both the consistency mask and the co-visibility mask improve performance, the consistency mask, which suppresses generation errors, has a greater impact. These results suggest that a well-designed weighting scheme that effectively filters out generation errors is key to improving performance.

\begin{figure*}[t]
    \centering
    \begin{minipage}[t]{0.19\linewidth}
        \centering
        \includegraphics[width=\linewidth]{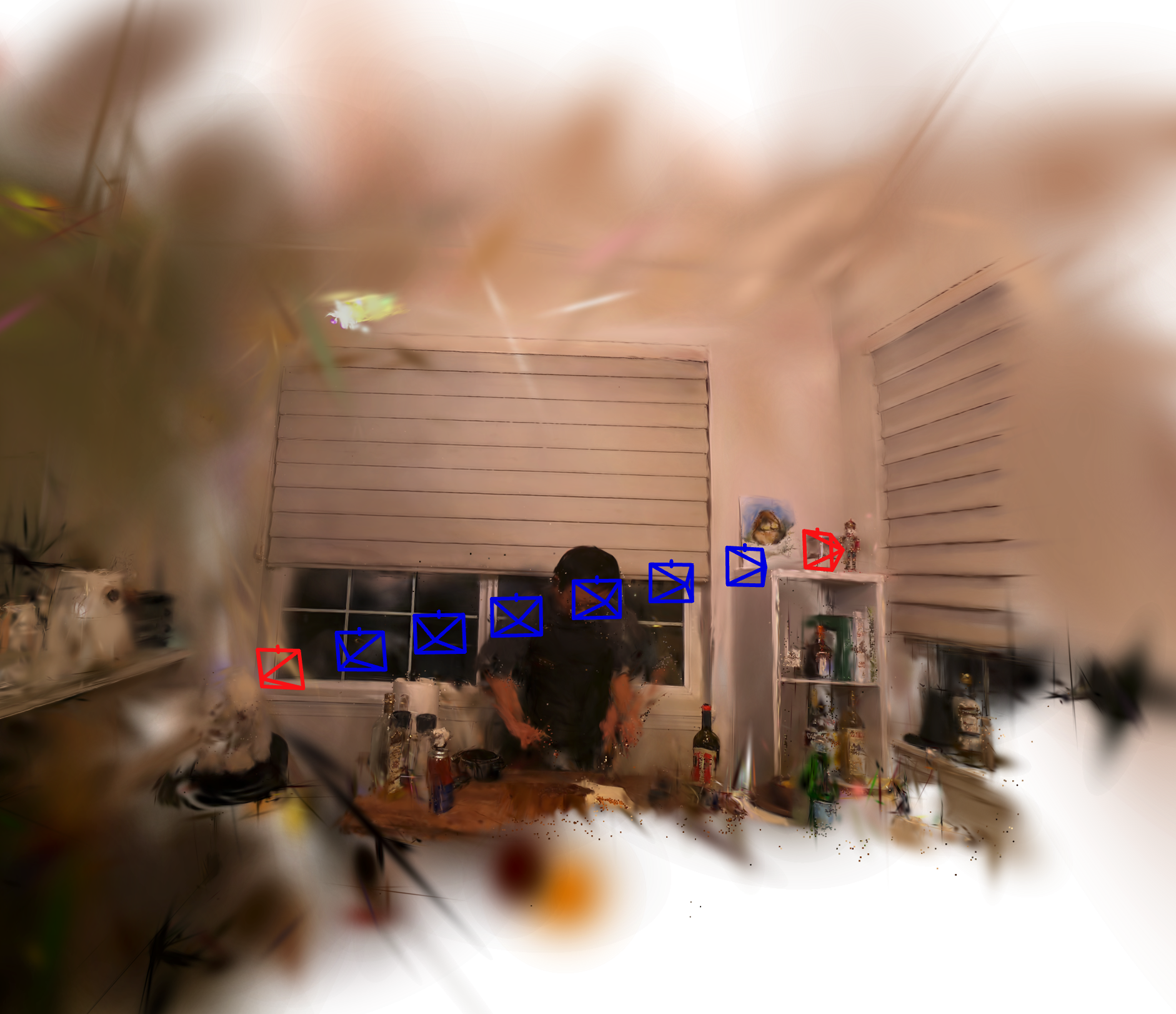}
        {\scriptsize{Interpolation~\cite{kong2025gsgs}}}\hfill
    \end{minipage}
    \begin{minipage}[t]{0.19\linewidth}
        \centering
        \includegraphics[width=\linewidth]{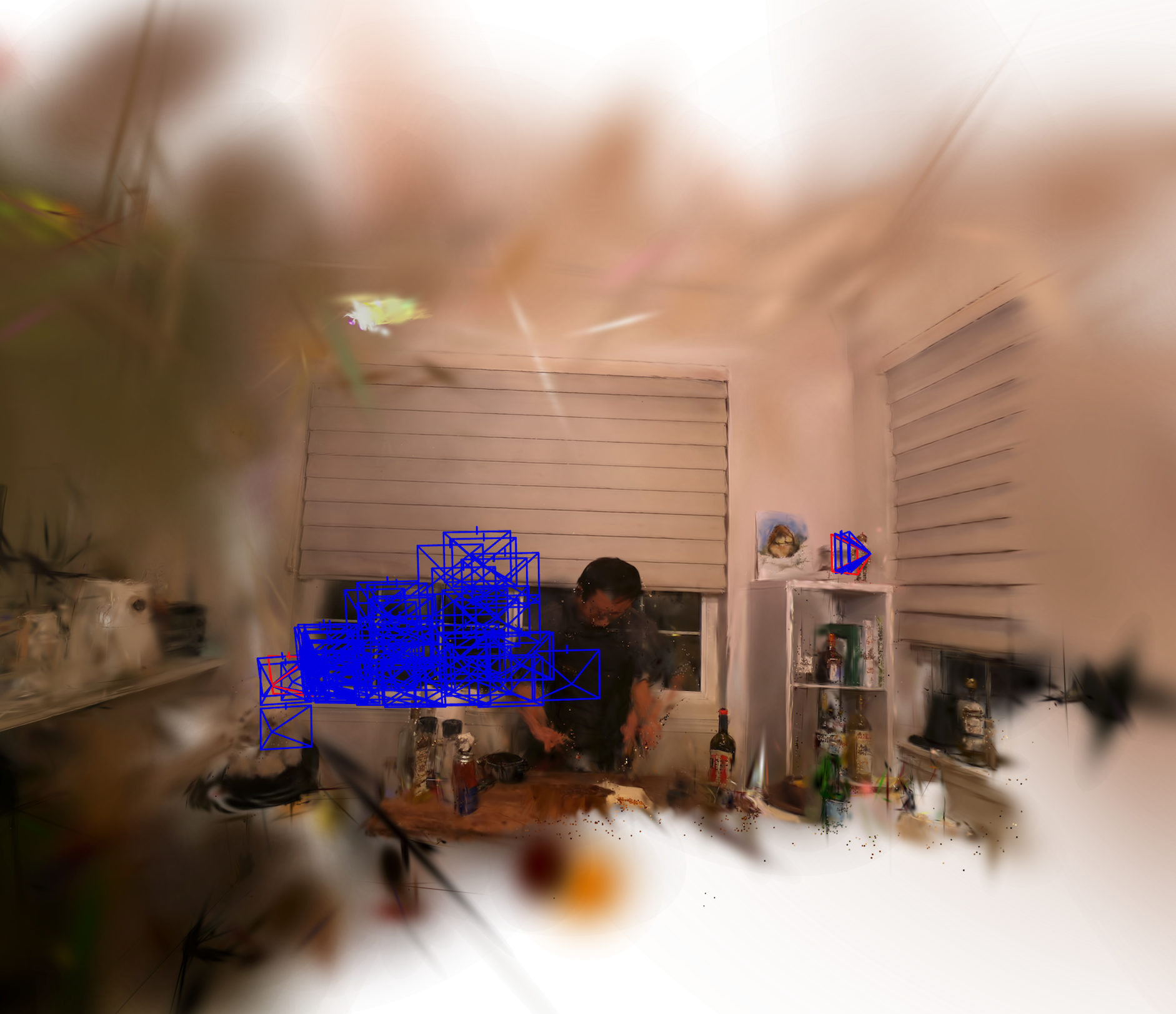}
        {\scriptsize{FisherRF~\cite{Jiang2023FisherRF}}}\hfill
    \end{minipage}
    \begin{minipage}[t]{0.19\linewidth}
        \centering
        \includegraphics[width=\linewidth]{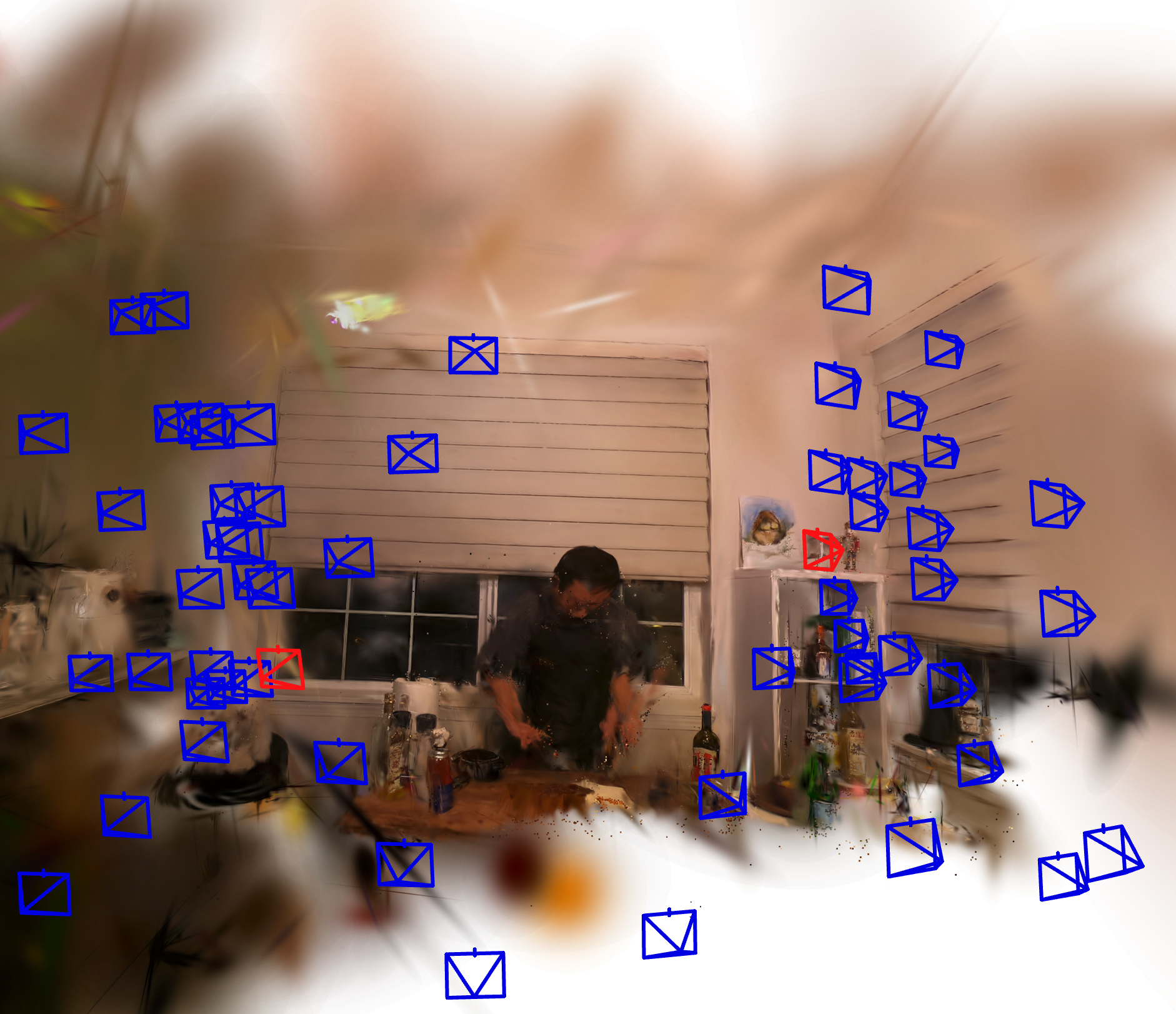}
        {\scriptsize{Coverage (3D)~\cite{kim2025exploregsexplorable3dscene}}}\hfill
    \end{minipage}
    \begin{minipage}[t]{0.19\linewidth}
        \centering
        \includegraphics[width=\linewidth]{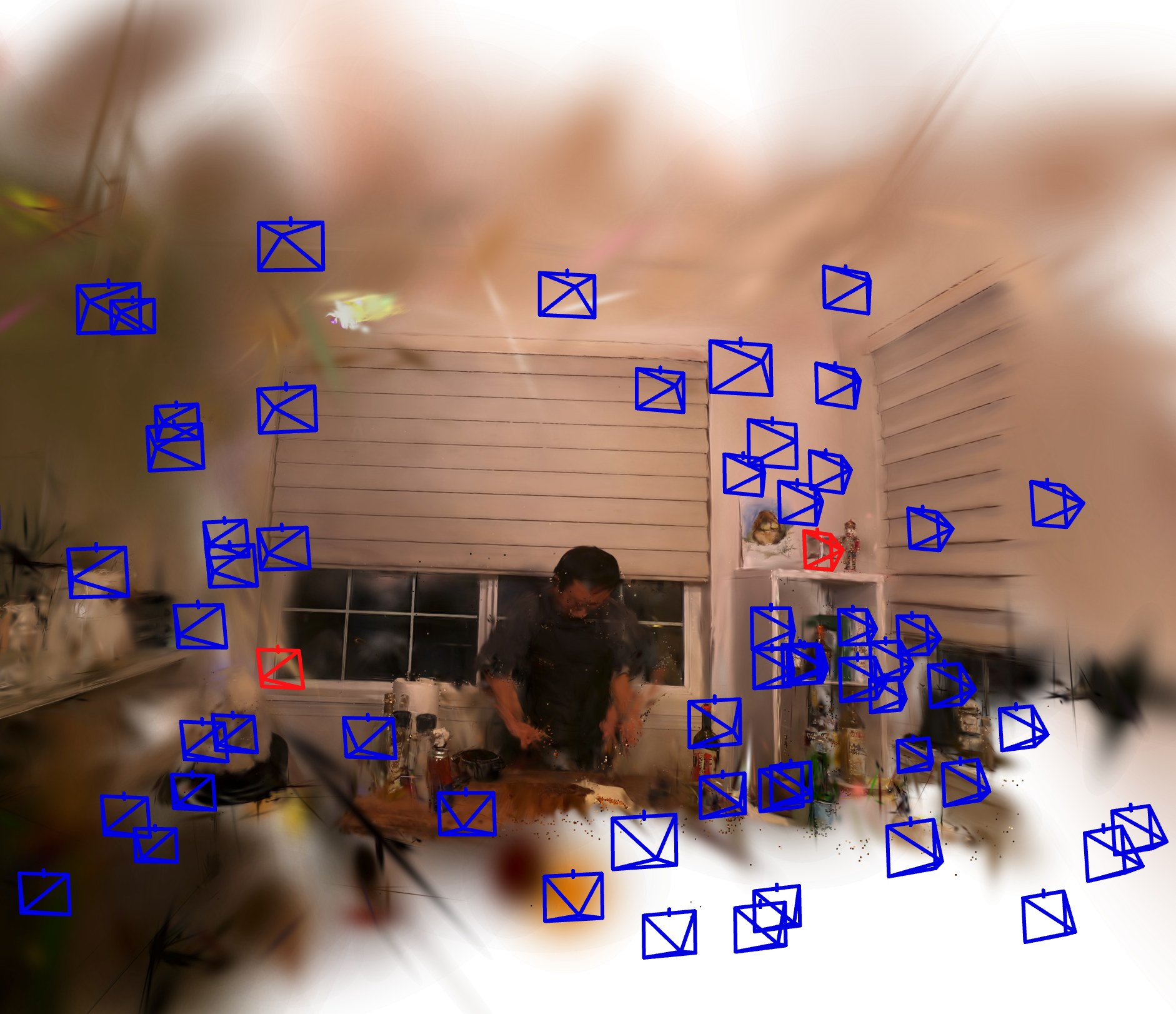}
        {\scriptsize{Coverage (4D)~\cite{kim2025exploregsexplorable3dscene}}}\hfill
    \end{minipage}
    \begin{minipage}[t]{0.19\linewidth}
        \centering
        \includegraphics[width=\linewidth]{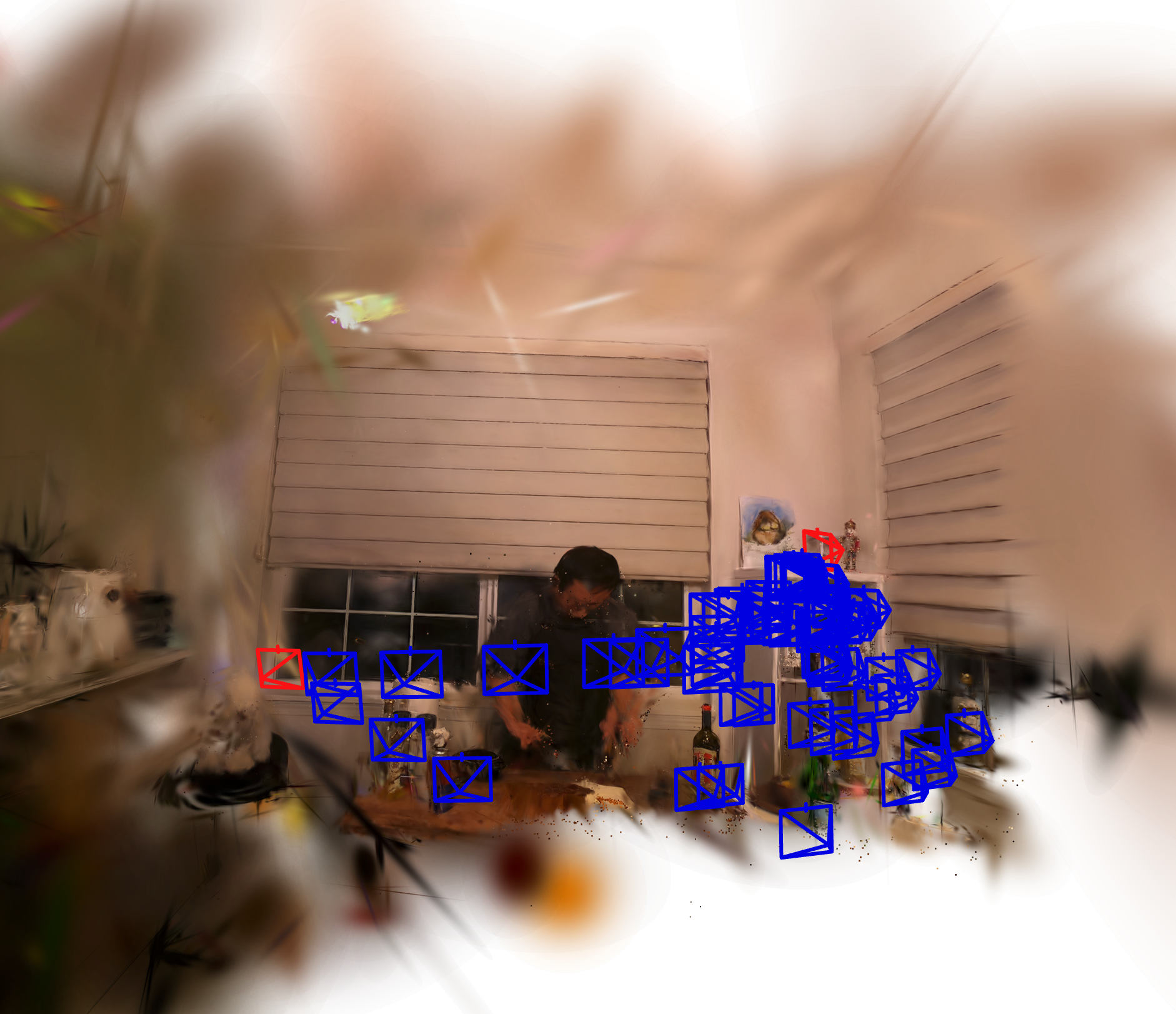}
        {\scriptsize{Ours}}\hfill
    \end{minipage}
    \begin{minipage}[t]{\linewidth}\centering{(a) Interpolation setting}\end{minipage}
    \begin{minipage}[t]{0.19\linewidth}
        \centering
        \includegraphics[width=\linewidth]{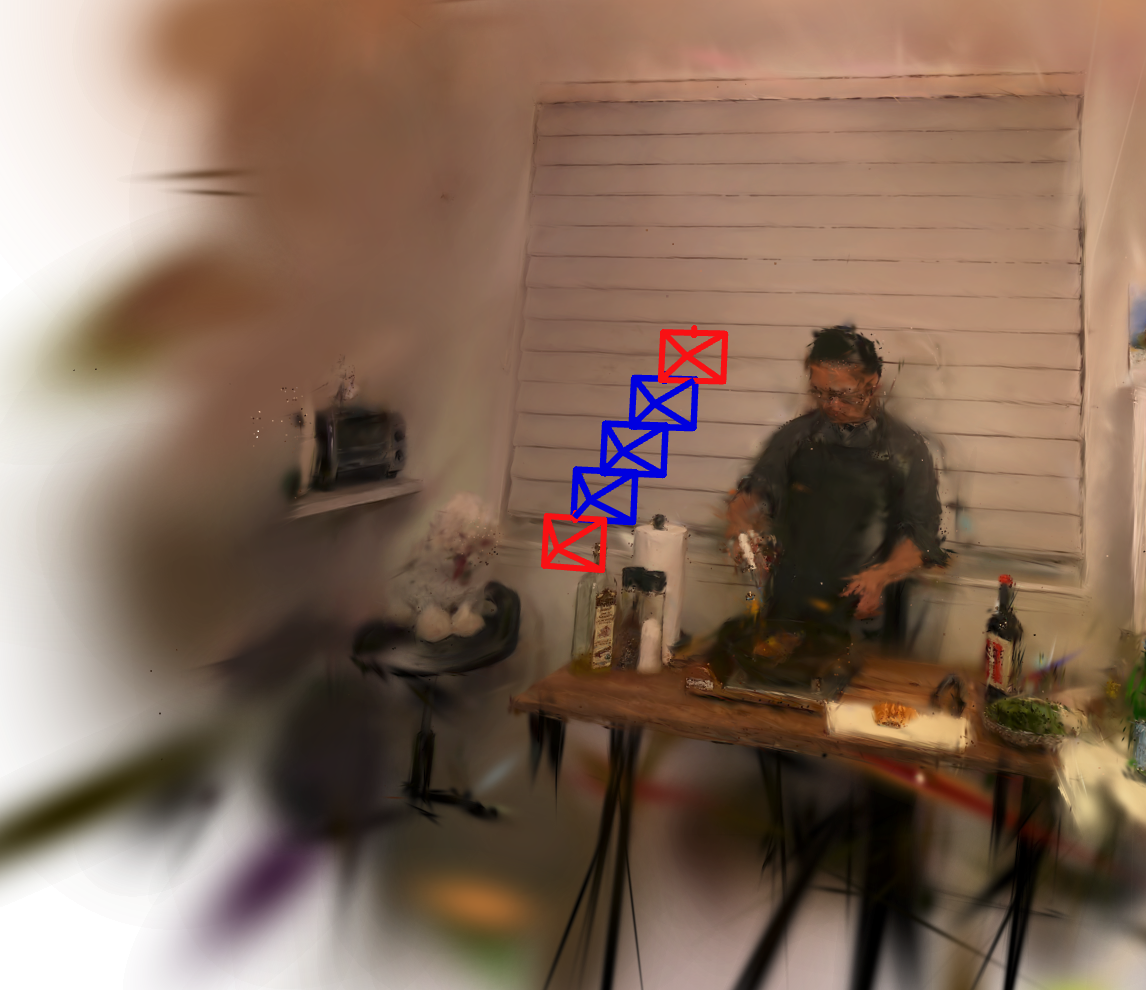}
        {\scriptsize{Interpolation~\cite{kong2025gsgs}}}\vspace{1mm}\hfill
    \end{minipage}
    \begin{minipage}[t]{0.19\linewidth}
        \centering
        \includegraphics[width=\linewidth]{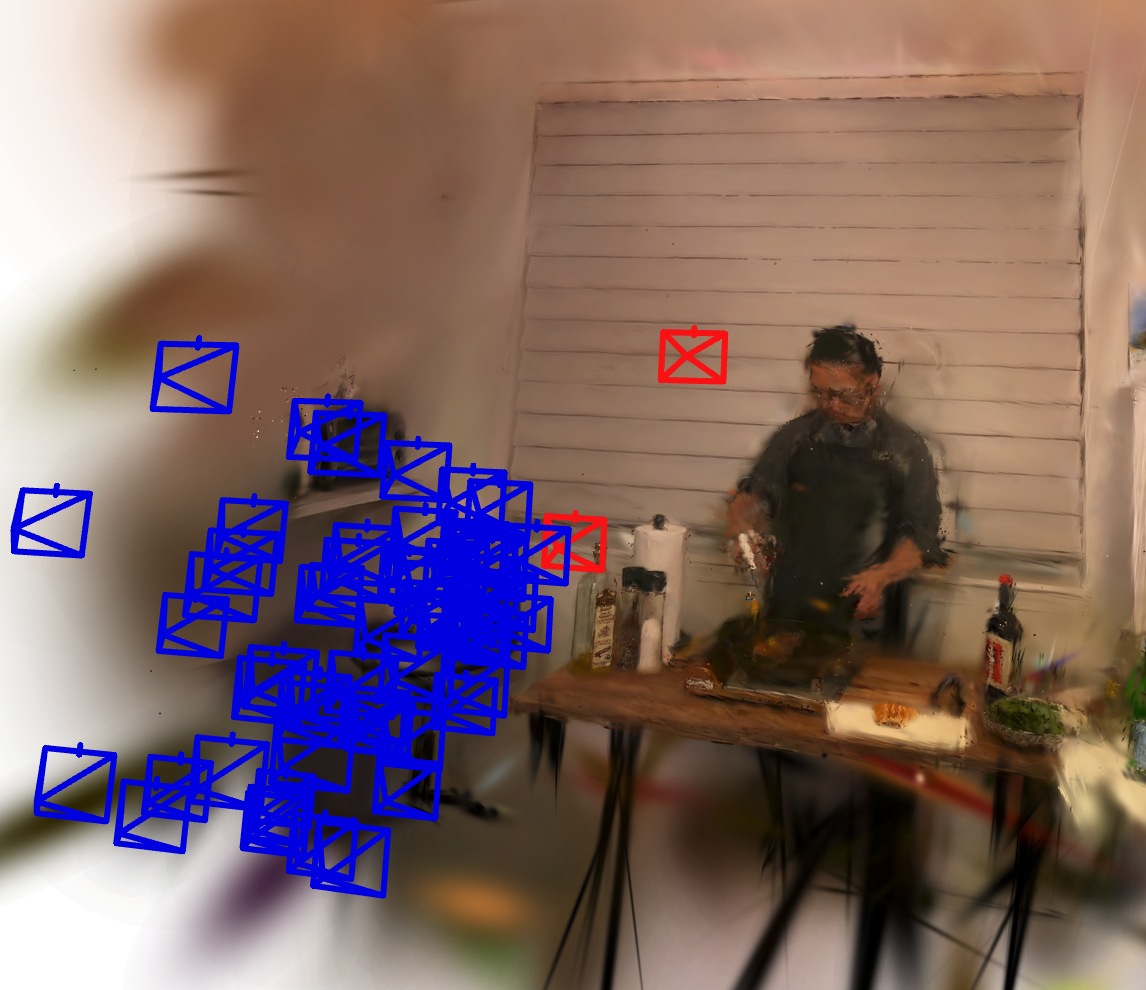}
        {\scriptsize{FisherRF~\cite{Jiang2023FisherRF}}}\vspace{1mm}\hfill
    \end{minipage}
    \begin{minipage}[t]{0.19\linewidth}
        \centering
        \includegraphics[width=\linewidth]{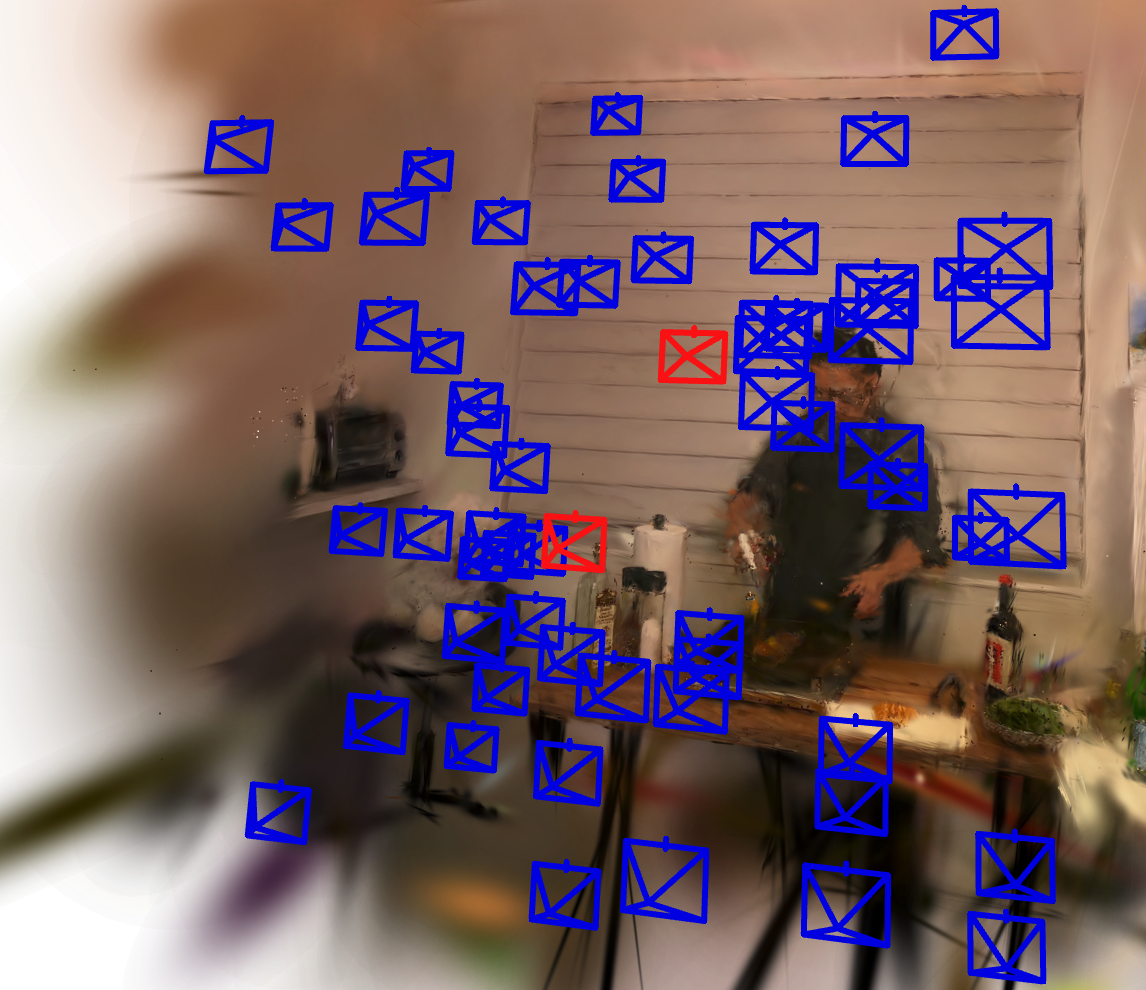}
        {\scriptsize{Coverage (3D)~\cite{kim2025exploregsexplorable3dscene}}}\vspace{1mm}\hfill
    \end{minipage}
    \begin{minipage}[t]{0.19\linewidth}
        \centering
        \includegraphics[width=\linewidth]{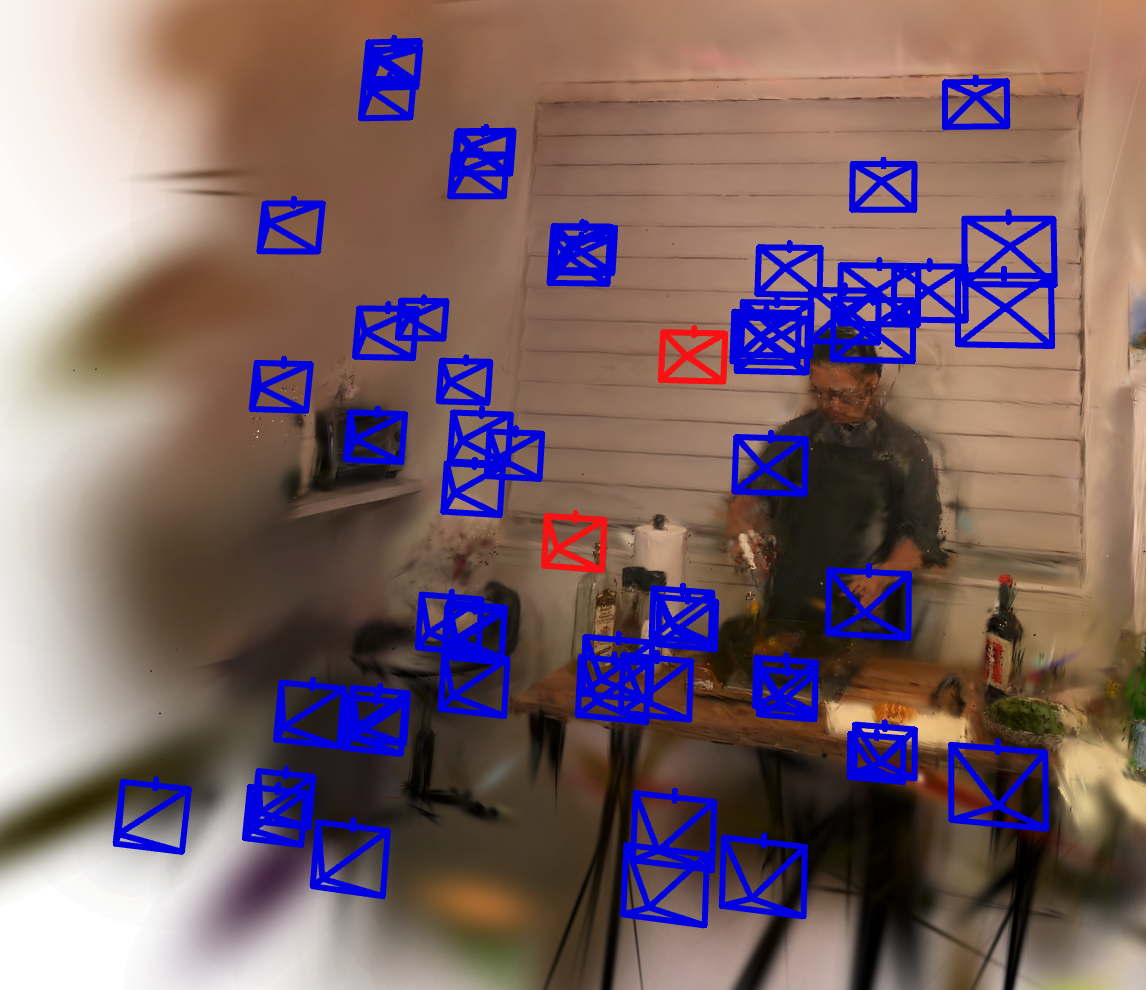}
        {\scriptsize{Coverage (4D)~\cite{kim2025exploregsexplorable3dscene}}}\vspace{1mm}\hfill
    \end{minipage}
    \begin{minipage}[t]{0.19\linewidth}
        \centering
        \includegraphics[width=\linewidth]{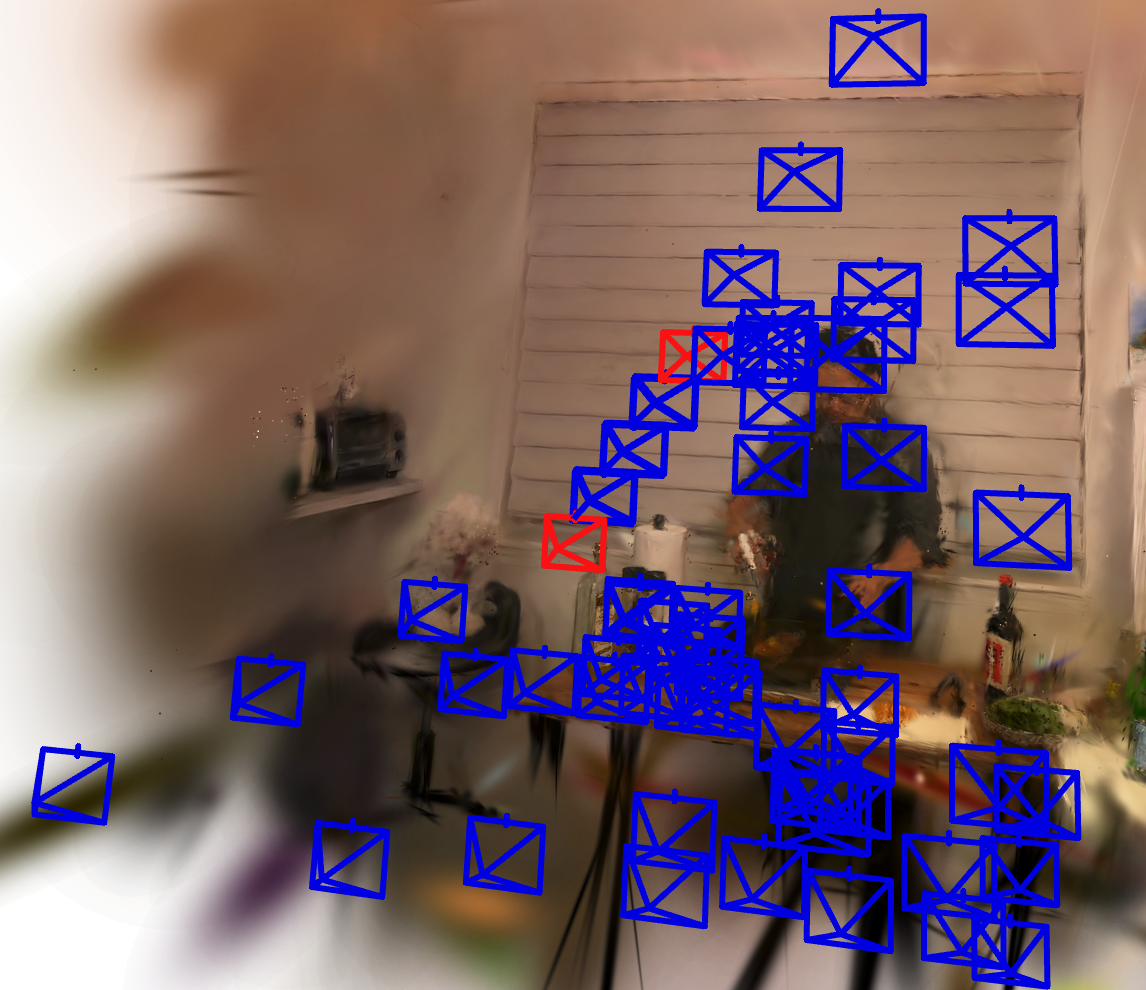}
        {\scriptsize{Ours}}\vspace{1mm}\hfill
    \end{minipage}
    \begin{minipage}[t]{\linewidth}\centering{(b) Extrapolation setting}\end{minipage}
    \vspace{-0.6cm}\caption{\textbf{Visualization of virtual trajectory sampling.} Red cameras indicate training cameras, and blue cameras indicate sampled virtual cameras.}
    \label{fig:vis_sampling}
\end{figure*}

\section{Conclusion}
We present a framework for active spatiotemporal virtual-viewpoint selection in 4D Gaussian Splatting. Our framework integrates (i) an active spatiotemporal virtual-viewpoint selection strategy to refine rendered images using rendering sensitivity and motion-aware observation density, and (ii) a fine-tuning method that mitigates generation errors and conflicts with real observations. We evaluate our method on multi-view video datasets using new train/test splits and show that it consistently outperforms existing viewpoint selection and fine-tuning approaches, reducing artifacts and improving rendering quality both quantitatively and qualitatively.

\noindent\textbf{Limitations.}
The representational capacity of the underlying 4DGS model is limited. Consequently, even with high-quality generated images, 4DGS may struggle to capture complex motion, rendering some regions unrecoverable. Importantly, this limitation is independent of our pipeline for integrating generated images, and may be mitigated by introducing additional constraints.

Moreover, our method is limited by the diffusion model’s generation quality. Sampling viewpoints that the model cannot reliably generate is computationally wasteful. Therefore, the sampling range should be carefully constrained. As diffusion models improve, this limitation may be alleviated.

\section*{Acknowledgements}
This work was partially supported by JST ACT-X (JPMJAX25C5), JST SPRING (Grant No.~JPMJSP2108), and JST ASPIRE (Grant No.~JPMJAP2303).

\bibliographystyle{splncs04}
\bibliography{main}

\newpage
\appendix

\maketitlesupplementary

\section{Additional Implementation Details}
\noindent\textbf{Video diffusion model.}
For training the video diffusion model, we use 1,000 static scenes from the DL3DV dataset~\cite{dl3dv}, following the same procedure as in 3DGS-Enhancer~\cite{liu20243dgs}. Similarly, for dynamic scenes, we construct a dataset by subsampling the input images to create artifact-prone settings. We use 50 scenes from DAVIS~\cite{davis}, 6 from the SelfCap dataset~\cite{selfcap}, 2 from the MobileStage dataset~\cite{mobilestage}, 4 from the HyperNeRF dataset~\cite{park2021hypernerf}, and 7 from the iPhone dataset~\cite{iphone}. The input image sequence is ordered in a serpentine manner with respect to the viewpoint index $x$ for each time step $t$, so that consecutive samples are adjacent in space or time. Formally, we define
\begin{equation}
x_t(k)=
\begin{cases}
k & (t\ \mathrm{even})\\
3-k & (t\ \mathrm{odd})
\end{cases}
\quad (k=0,\ldots,3).
\end{equation}
We then concatenate the samples in increasing order of $t$ ($t=0,1,2$), scanning $k=0,\ldots,3$ within each time step. For a static scene or a monocular video, training is carried out by masking all input and output regions except for a target region. We use a single NVIDIA A100 80 GB GPU to train the U-Net and the VAE decoder. Training takes approximately 2 days for the U-Net and 14 days for the VAE decoder.

\noindent\textbf{Exploration space.}
We follow ExploreGS~\cite{kim2025exploregsexplorable3dscene} to define the exploration space. We first construct a bounding box based on the input images and the reconstruction model. We then discretize the bounding box into a uniform grid with 64 divisions along each axis. We mask out voxels corresponding to occupied regions and do not sample viewpoint candidates within occupied voxels. We set the step size for generating virtual viewpoint candidates to twice the body-diagonal length of a single grid cell.

\begin{table*}[t]
    \centering
    \small
    \begin{adjustbox}{width=1.0\linewidth}
    \begin{tabular}{c|rrrrr|rrrrr}
    \renewcommand{\arraystretch}{1.05}
    & \multicolumn{5}{c|}{1 trajectory} & \multicolumn{5}{c}{5 trajectories}\\
    & PSNR$\uparrow$ & SSIM$\uparrow$ & LPIPS$\downarrow$ & FID$\downarrow$ & DINOv2$\uparrow$
    & PSNR$\uparrow$ & SSIM$\uparrow$ & LPIPS$\downarrow$ & FID$\downarrow$ & DINOv2$\uparrow$\\\hline
    Interpolation-based selection~\cite{kong2025gsgs} & 20.67 & 0.774 & 0.296 & 121.28 & 0.889 & 20.73 & 0.776 & 0.294 & 117.60 & 0.892 \\
    FisherRF-based selection~\cite{Jiang2023FisherRF} & \textbf{21.31} & 0.777 & 0.305 & 138.46 & 0.852 & 20.39 & 0.768 & 0.306 & 132.19 & 0.854 \\
    Coverage-based (3D) selection~\cite{kim2025exploregsexplorable3dscene} & 19.76 & 0.748 & 0.326 & 152.64 & 0.853 & 20.18 & 0.763 & 0.311 & 127.51 & 0.891 \\
    Coverage-based (4D) selection~\cite{kim2025exploregsexplorable3dscene} & 19.82 & 0.753 & 0.327 & 160.68 & 0.869 & 20.49 & 0.764 & 0.312 & 141.46 & 0.888 \\
    Ours & 20.89 & \textbf{0.780} & \textbf{0.286} & \textbf{105.57} & \textbf{0.903} & \textbf{21.69} & \textbf{0.784} & \textbf{0.269} & \textbf{103.97} & \textbf{0.912} \\
    \end{tabular}
    \end{adjustbox}

    \vspace{1mm}

    \begin{adjustbox}{width=1.0\linewidth}
    \begin{tabular}{c|rrrrr|rrrrr}
    \renewcommand{\arraystretch}{1.05}
    & \multicolumn{5}{c|}{10 trajectories} & \multicolumn{5}{c}{20 trajectories}\\
    & PSNR$\uparrow$ & SSIM$\uparrow$ & LPIPS$\downarrow$ & FID$\downarrow$ & DINOv2$\uparrow$
    & PSNR$\uparrow$ & SSIM$\uparrow$ & LPIPS$\downarrow$ & FID$\downarrow$ & DINOv2$\uparrow$\\\hline
    Interpolation-based selection~\cite{kong2025gsgs} & 20.72 & 0.776 & 0.293 & 118.93 & 0.896 & 20.75 & 0.777 & 0.293 & 120.08 & 0.896 \\
    FisherRF-based selection~\cite{Jiang2023FisherRF} & 21.00 & 0.773 & 0.306 & 132.87 & 0.866 & 21.01 & 0.774 & 0.303 & 135.36 & 0.853 \\
    Coverage-based (3D) selection~\cite{kim2025exploregsexplorable3dscene} & 20.70 & 0.771 & 0.298 & 131.35 & 0.904 & 19.91 & 0.754 & 0.317 & 148.68 & 0.878 \\
    Coverage-based (4D) selection~\cite{kim2025exploregsexplorable3dscene} & 19.97 & 0.756 & 0.323 & 139.16 & 0.882 & 19.09 & 0.737 & 0.325 & 140.19 & 0.866 \\
    Ours & \textbf{21.90} & \textbf{0.786} & \textbf{0.267} & \textbf{104.55} & \textbf{0.914} & \textbf{21.73} & \textbf{0.792} & \textbf{0.256} & \textbf{83.81} & \textbf{0.920} \\
    \end{tabular}
    \end{adjustbox}

    \caption{\textbf{Quantitative comparison of the number of trajectories on the Neural 3D Video Dataset~\cite{dynerf}.}}
    \label{table:num_traj_sensitivity}
\end{table*}

\noindent\textbf{Calculation of Gaussian speeds and observation counts.}
We define the speed of Gaussian $g_i$ at time $t$, denoted by $s_{g_i}(t)$, using a central difference approximation:
\begin{equation}
s_{g_i}(t)
=
\left\|
\frac{\mathbf{p}_{g_i}(t+\Delta t)-\mathbf{p}_{g_i}(t-\Delta t)}{2\Delta t}
\right\|_2,
\label{eq:gaussian_velocity}
\end{equation}
where $\mathbf{p}_{g_i}(t)\in\mathbb{R}^3$ denotes the 3D position of Gaussian $g_i$ at time $t$, and $\Delta t$ is the time interval between consecutive observed frames. Visibility is determined as in 3D Gaussian Splatting~\cite{kerbl3Dgaussians}. Each Gaussian deemed visible is counted as one observation. The resulting speed $s_{g_i}(t)$ and the corresponding observation count are used to compute the motion-aware observation deficiency.

\noindent\textbf{Calculation of rendering sensitivity.}
We evaluate the rendering sensitivity of each candidate view by applying small perturbations along the translation, rotation, and temporal dimensions. For translation, we set the perturbation magnitude to one-fifth of the spatial voxel size and construct three perturbations along the \(x\)-, \(y\)-, and \(z\)-axes. For rotation, we use \(0.2^\circ\) as the perturbation magnitude and construct three rotational perturbations around the corresponding coordinate axes. For the temporal dimension, we use the frame interval of the input video as the perturbation magnitude. For each dimension, the sensitivity is defined as the squared error between the rendered images obtained from symmetric positive and negative perturbations. The resulting sensitivity scores for translation, rotation, and time are then normalized independently across candidate views before aggregation. This normalization mitigates the effect of dimension-specific absolute scale differences while preserving the relative sensitivity ordering within each dimension.

\noindent\textbf{Local strategy of candidate-viewpoint creation.}
Candidate viewpoint creation in the local strategy follows ExploreGS~\cite{kim2025exploregsexplorable3dscene}. For camera translation, we use six translation candidates obtained by translating along each axis in the positive and negative directions with a fixed step size. For camera rotation, we construct four rotation candidates by applying a fixed angular step of \(10^\circ\) in the positive and negative directions about the two axes orthogonal to the camera’s forward direction.

\section{More Results}
\subsection{Sensitivity to the Number of Virtual Trajectories}
\Tref{table:num_traj_sensitivity} shows the results for different numbers of virtual trajectories. The proposed method shows only a minor degradation even with a small number of trajectories (1 or 5 trajectories), indicating high sample efficiency under a limited sampling budget. In contrast, the baseline methods degrade substantially when fewer trajectories are available and achieve only marginal improvements as the number of trajectories increases. In particular, the coverage-based selection exhibits unstable behavior as the number of trajectories varies, failing to yield consistent gains. These trends suggest that simply increasing the number of trajectories does not necessarily lead to more informative observations, whereas the proposed method utilizes the trajectory budget more effectively by prioritizing informative viewpoints. As a result, the proposed method remains robust and maintains high performance even when the number of trajectories is constrained.

\begin{table*}[t]
    \centering
    \begin{adjustbox}{width=1.0\linewidth}
    \small\begin{tabular}{c|rrrrr|c}
    \renewcommand{\arraystretch}{1.05}
    & PSNR$\uparrow$ & SSIM$\uparrow$ & LPIPS$\downarrow$ & FID$\downarrow$ & DINOv2$\uparrow$ & Generation cost \\\hline
    E-D3DGS~\cite{bae2024ed3dgs} & 20.42 & 0.775 & 0.293 & 129.33 & 0.871 & - \\
    Interpolation-based selection~\cite{kong2025gsgs} & 20.75 & 0.777 & 0.293 & 120.08 & 0.896 & 14 min \\
    FisherRF-based selection~\cite{Jiang2023FisherRF} & 21.01 & 0.774 & 0.303 & 135.36 & 0.853 & 15 min \\
    Coverage-based (3D) selection~\cite{kim2025exploregsexplorable3dscene} & 19.91 & 0.754 & 0.317 & 148.68 & 0.878 & 14 min \\
    Coverage-based (4D) selection~\cite{kim2025exploregsexplorable3dscene} & 19.09 & 0.737 & 0.325 & 140.19 & 0.866 & 14 min \\
    Ours (10 trajectories) & \textbf{21.90} & 0.786 & 0.267 & 104.55 & 0.914 & 13 min \\
    Ours (20 trajectories) & 21.73 & \textbf{0.792} & \textbf{0.256} & \textbf{83.81} & \textbf{0.920} & 20 min \\
    \end{tabular}
    \end{adjustbox}
    \caption{\textbf{Analysis of the computational cost of the generation pipeline.} Although our approach is more computationally expensive than existing methods, it still achieves better performance than the baselines even within a comparable time budget.}
    \label{table:generation_cost}
\end{table*}

\begin{figure*}[!t]
    \centering
    \setlength{\tabcolsep}{0.02cm}
    \setlength{\itemwidth}{2.4cm}
    \renewcommand{\arraystretch}{0.5}
    \hspace*{-\tabcolsep}\small\begin{tabular}{cc}
            \includegraphics[width=0.4\linewidth]{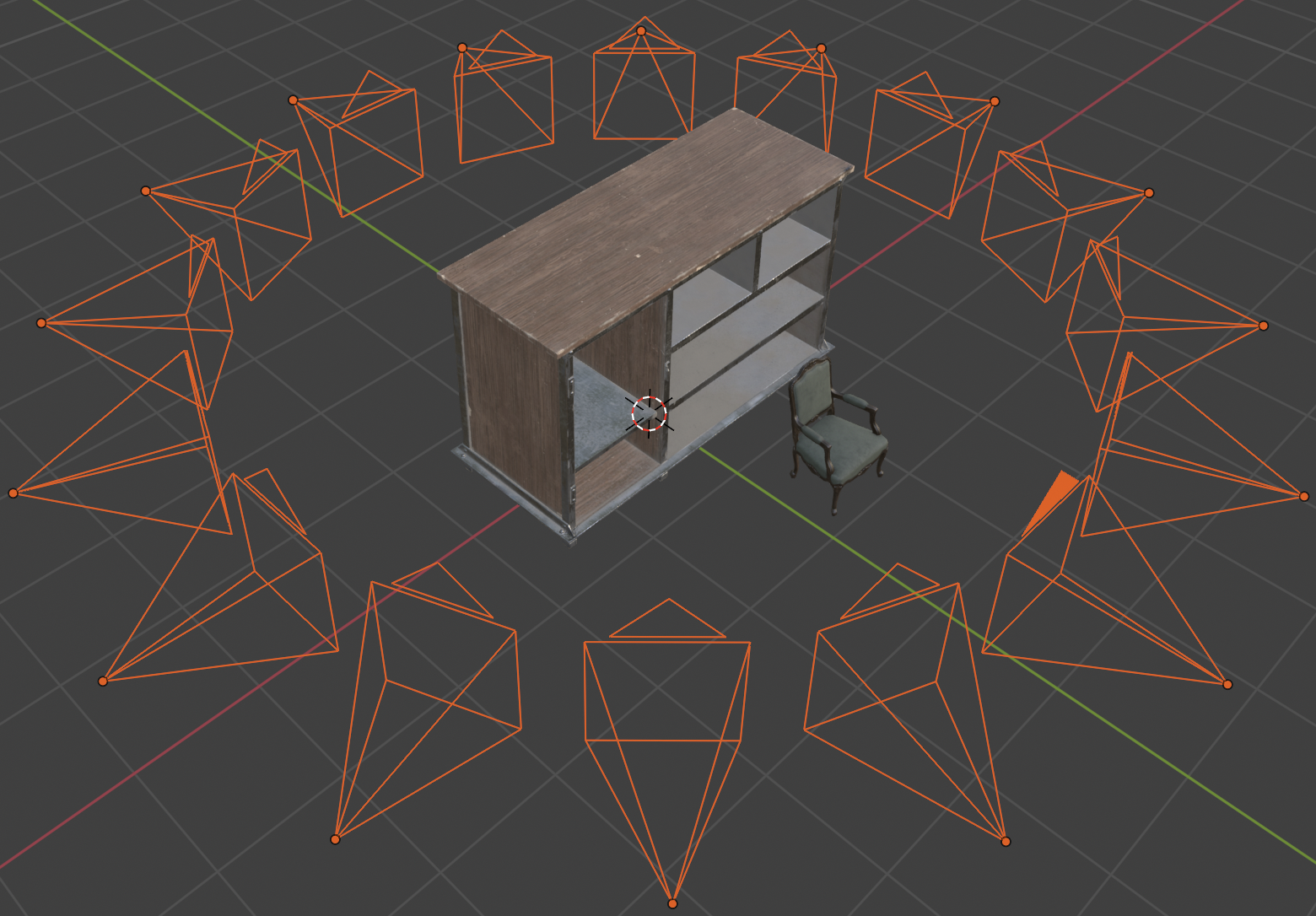} &
            \includegraphics[width=0.3\linewidth]{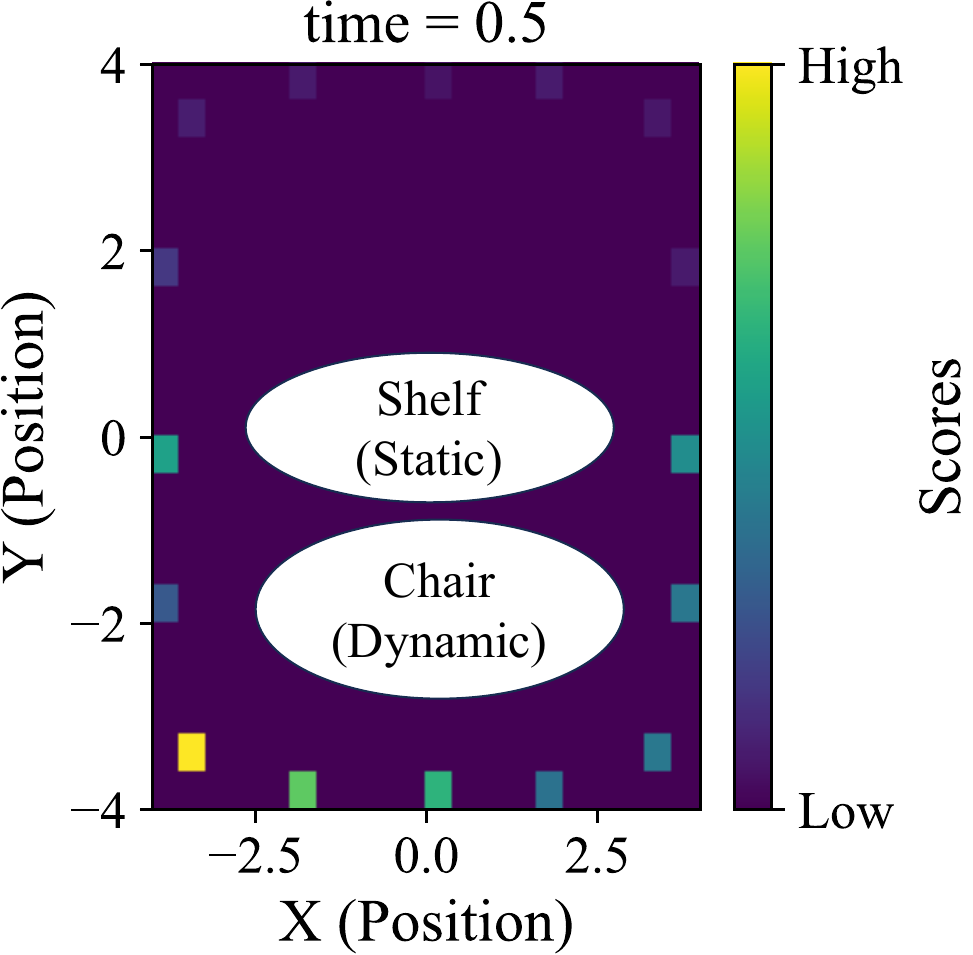} \\
            \vspace{0.2em}
            (a) Scene setup &
            (b) Visualizing scores over space
        \\
    \end{tabular}
    \\
    \includegraphics[width=0.65\linewidth]{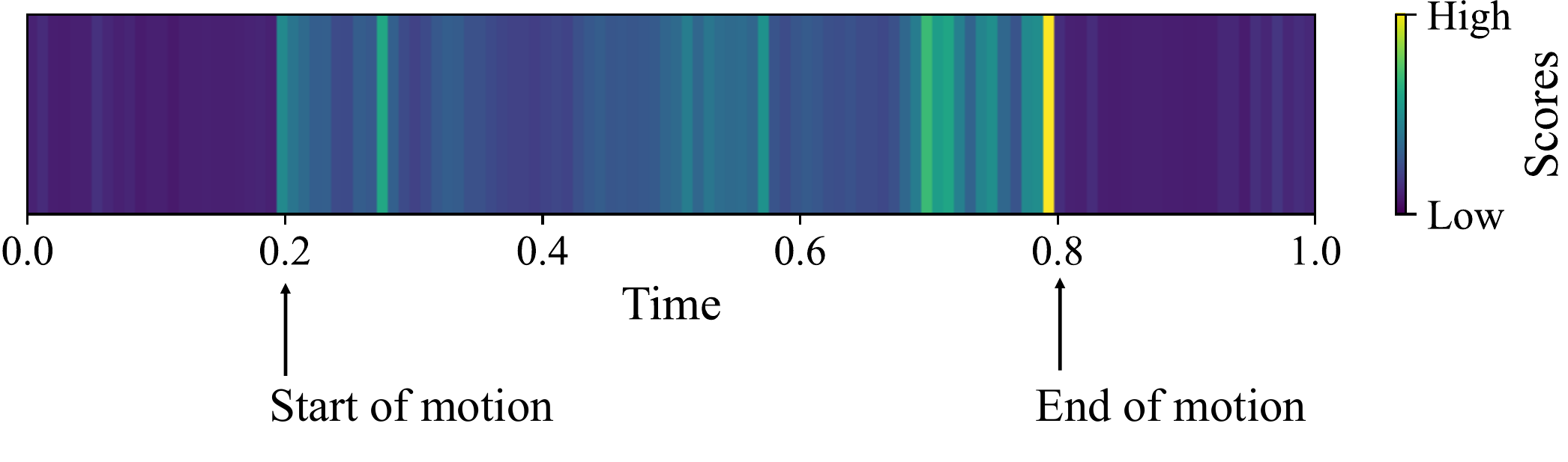}
    \vspace{0.1cm}\text{(c) Visualizing scores over time.}
    \begin{adjustbox}{width=1.0\linewidth}
    \hspace*{-\tabcolsep}\small\begin{tabular}{cccccc}
            \includegraphics[width=\itemwidth]{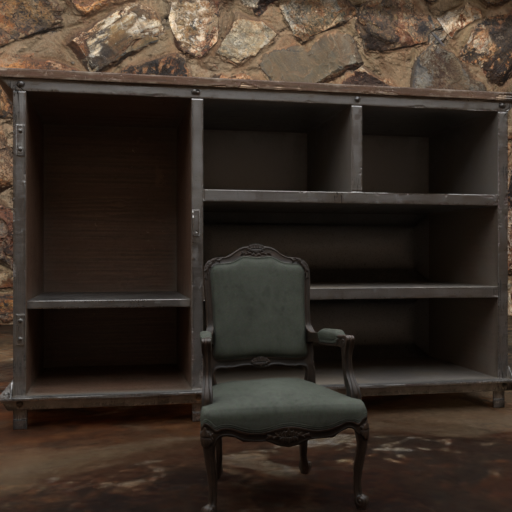} &
            \includegraphics[width=\itemwidth]{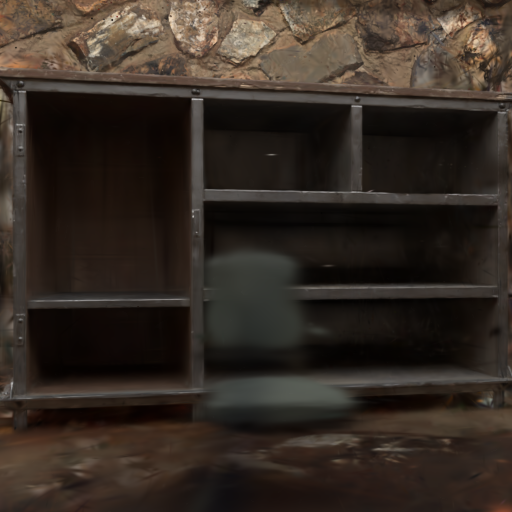} &
            \includegraphics[width=\itemwidth]{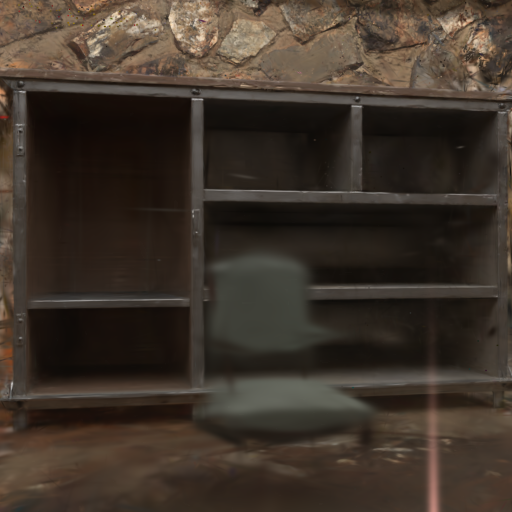} &
            \includegraphics[width=\itemwidth]{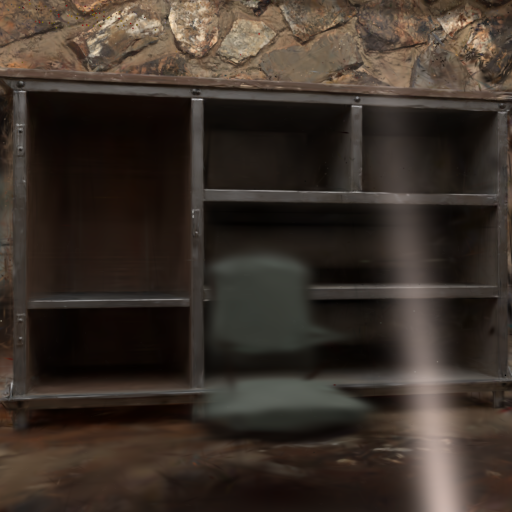} &
            \includegraphics[width=\itemwidth]{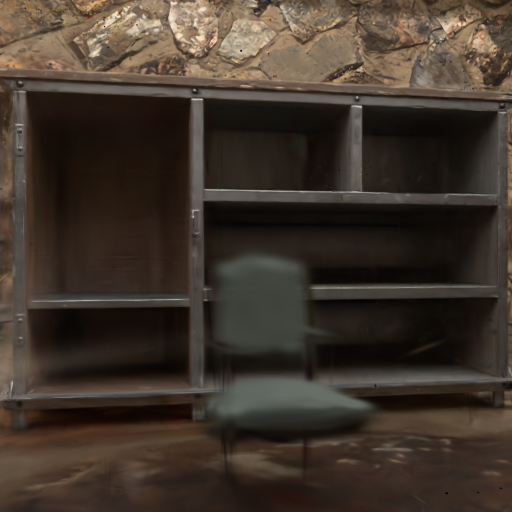} &
            \includegraphics[width=\itemwidth]{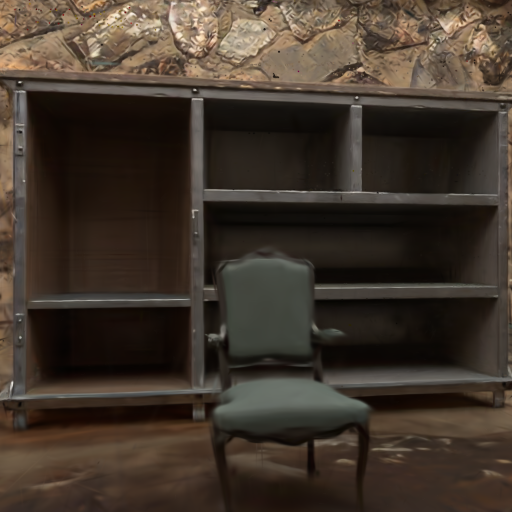} \\
            \capcell{Ground-Truth} &
            \capcell{E-D3DGS~\cite{bae2024ed3dgs}} &
            \capcell{Interpolation-based\\selection~\cite{kong2025gsgs}} &
            \capcell{Coverage-based\\(3D) selection~\cite{kim2025exploregsexplorable3dscene}} &
            \capcell{Coverage-based\\(4D) selection~\cite{kim2025exploregsexplorable3dscene}} &
            \capcell{Ours} \\
        \\
    \end{tabular}\vspace{-1.5em}
    \end{adjustbox}
    \vspace{0.2cm}\text{(d) Novel views.}
    \vspace{-0.4cm}\caption{\textbf{Evaluation of a starting-viewpoint selection strategy using a synthetic dataset generated in Blender.} (a) Scene setup without floors or walls for visualization. (b) Our approach samples starting viewpoints from regions relevant to dynamic objects. (c) Our approach focuses on time periods with motion. (d) Our approach achieves better reconstruction quality than existing methods.}
    \label{fig:syn_analysis}
\end{figure*}

\subsection{Computational Cost}
Because our viewpoint-selection method requires multiple renderings, it introduces additional computation during the selection stage. However, as shown in \Tref{table:generation_cost}, the proposed method remains competitive even under a low-cost setting using fewer trajectories: with 10 trajectories, it achieves a lower end-to-end generation cost than the 20-trajectory baseline while still delivering better performance. All results were measured with generation batch size 1 at 512$\times$512 resolution. Overall, these results suggest that the additional computation can be incorporated with negligible runtime overhead, yielding a favorable trade-off between efficiency and performance.

\subsection{Effects of Starting-Viewpoint Selection Strategy with a Synthetic Dataset}
To further analyze the effectiveness of our starting-viewpoint selection strategy, we visualize spatiotemporal selection scores in a synthetic scene. The scene is constructed in Blender and contains 16 training cameras covering \(360^\circ\), placed around the center in \(22.5^\circ\) intervals (\Fref{fig:syn_analysis}(a)). A chair and a shelf are placed in the scene, and only the chair is moved. Because the chair is occluded by the shelf during motion, only a subset of cameras can observe it. With the time scale set to 1.0, the chair’s motion occurs only in the interval from 0.2 to 0.8. As a result, for certain spatiotemporal regions, no viewpoints require refinement. In contrast, static structures such as the shelf and walls are observed throughout the entire time range and therefore have lower priority as refinement targets. Under these conditions, we visualize the starting-viewpoint selection scores over viewpoints (\Fref{fig:syn_analysis}(b)) and over time (\Fref{fig:syn_analysis}(c)). The visualization confirms that the proposed method assigns high scores to times and viewpoints where the chair motion occurs and is observable. This indicates that the proposed method selects informative observed viewpoints for reconstruction. Moreover, \Fref{fig:syn_analysis}(d) shows that our proposed method improves rendering quality in novel views.

\begin{figure*}[t]
    \centering
    \setlength{\tabcolsep}{0.02cm}
    \setlength{\itemwidth}{2.4cm}
    \renewcommand{\arraystretch}{0.5}
    \begin{adjustbox}{width=1.0\linewidth}
    \hspace*{-\tabcolsep}\small\begin{tabular}{cccccc}
            \includegraphics[width=1.33333333333\itemwidth]{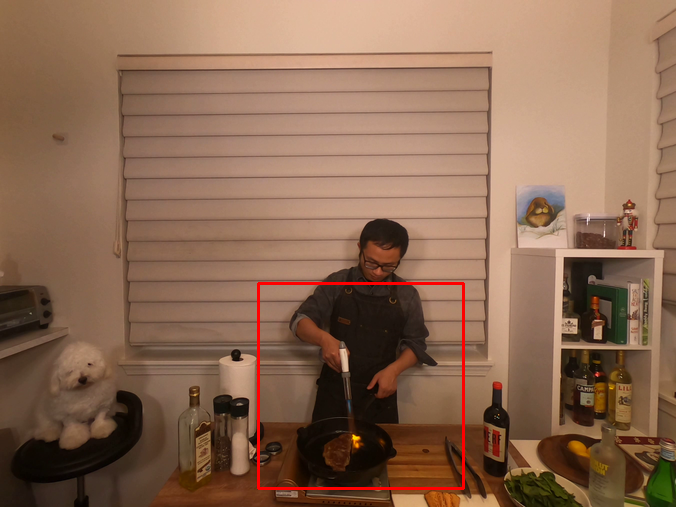} &
            \fboxsep=0pt\fcolorbox{red}{white}{\includegraphics[width=\itemwidth]{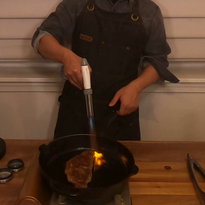}} &
            \includegraphics[width=\itemwidth]{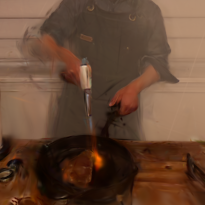} &
            \includegraphics[width=\itemwidth]{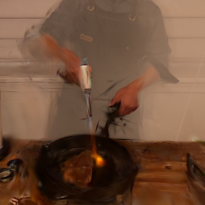} &
            \includegraphics[width=\itemwidth]{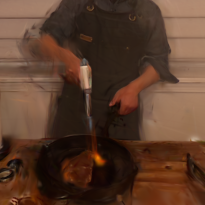} &
            \includegraphics[width=\itemwidth]{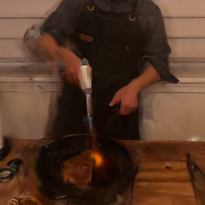} \\
            \includegraphics[width=1.33333333333\itemwidth]{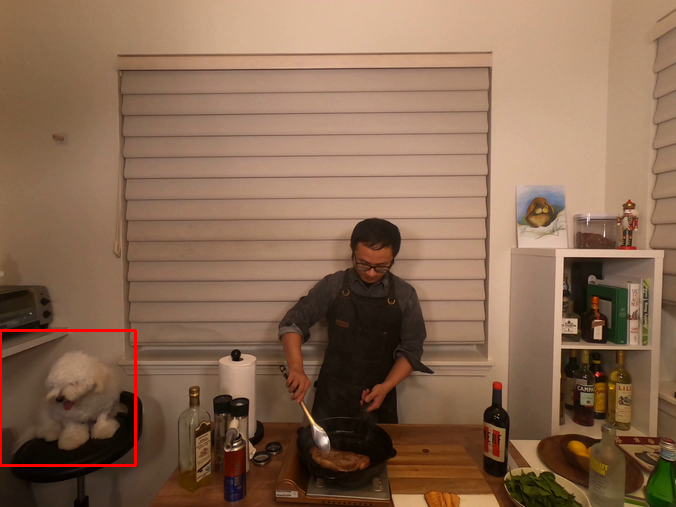} &
            \fboxsep=0pt\fcolorbox{red}{white}{\includegraphics[width=\itemwidth]{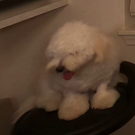}} &
            \includegraphics[width=\itemwidth]{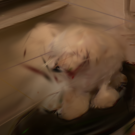} &
            \includegraphics[width=\itemwidth]{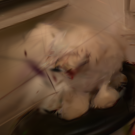} &
            \includegraphics[width=\itemwidth]{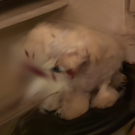} &
            \includegraphics[width=\itemwidth]{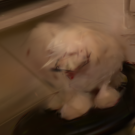} \\
            \multicolumn{2}{c}{Ground-Truth image} &
            \capcell{Ours w/o\\global candidate view creation} &
            \capcell{Ours w/o\\starting-viewpoint selection} &
            \capcell{Ours w/o\\candidate-viewpoint selection} &
            \capcell{Ours} \\
        \\
    \end{tabular}\vspace{-1.5em}
    \end{adjustbox}
  \vspace{-0.4cm}\caption{\textbf{Qualitative results of the ablation study on our viewpoint selection.}}
  \label{fig:ablation_selection}
\end{figure*}

\begin{figure*}[t]
    \centering
    \setlength{\tabcolsep}{0.02cm}
    \setlength{\itemwidth}{2.4cm}
    \renewcommand{\arraystretch}{0.5}
    \begin{adjustbox}{width=1.0\linewidth}
    \hspace*{-\tabcolsep}\small\begin{tabular}{cccccc}
            \includegraphics[width=1.33333333333\itemwidth]{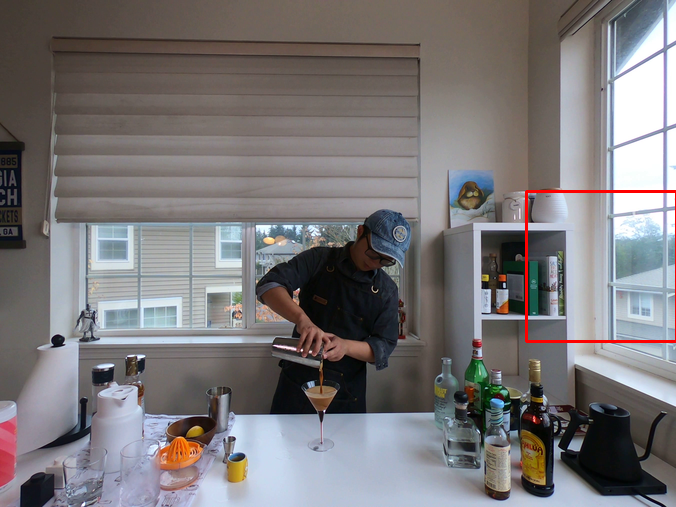} &
            \fboxsep=0pt\fcolorbox{red}{white}{\includegraphics[width=\itemwidth]{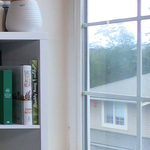}} &
            \includegraphics[width=\itemwidth]{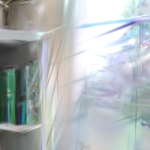} &
            \includegraphics[width=\itemwidth]{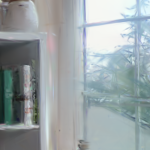} &
            \includegraphics[width=\itemwidth]{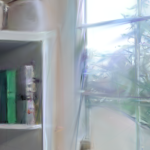} &
            \includegraphics[width=\itemwidth]{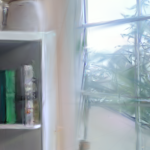} \\
            \includegraphics[width=1.33333333333\itemwidth]{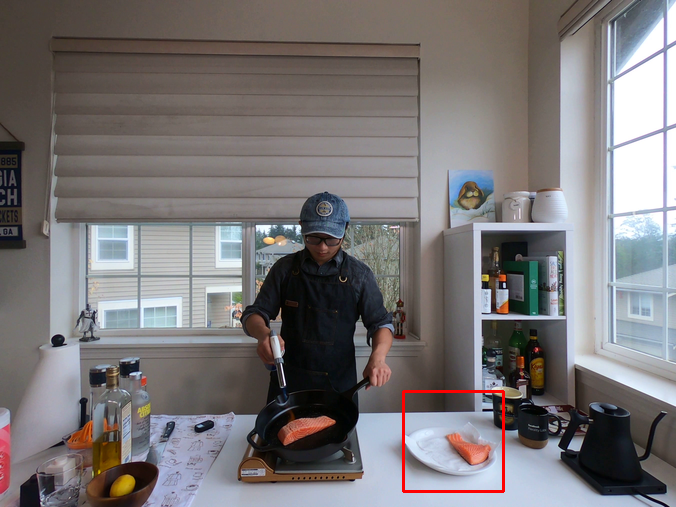} &
            \fboxsep=0pt\fcolorbox{red}{white}{\includegraphics[width=\itemwidth]{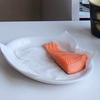}} &
            \includegraphics[width=\itemwidth]{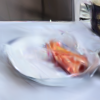} &
            \includegraphics[width=\itemwidth]{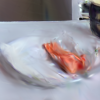} &
            \includegraphics[width=\itemwidth]{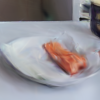} &
            \includegraphics[width=\itemwidth]{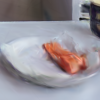} \\
            \multicolumn{2}{c}{Ground-Truth image} &
            \capcell{Ours w/o weight} &
            \capcell{Ours w/o\\consistency mask} &
            \capcell{Ours w/o\\co-visibility mask} &
            \capcell{Ours} \\
        \\
    \end{tabular}\vspace{-1.5em}
    \end{adjustbox}
  \vspace{-0.4cm}\caption{\textbf{Qualitative results of the ablation study on our fine-tuning methods.}}
  \label{fig:ablation_finetuning}
\end{figure*}

\subsection{Qualitative Comparison of Ablation Studies}
\noindent\textbf{Viewpoint selection.}
\Fref{fig:ablation_selection} shows the rendered results of an ablation study on our viewpoint selection. When components of our viewpoint selection method are removed, artifacts appear (top row), and the refinement is insufficient (bottom row). In contrast, our full model resolves these issues.

\noindent\textbf{Fine-tuning methods.}
\Fref{fig:ablation_finetuning} shows the rendered results of an ablation study on our fine-tuning methods. When the weighting scheme is removed or when background updates are not suppressed by co-visibility masks, the background regions degrade (top row). In addition, using the consistency mask, only regions consistent with the input image are selectively incorporated, resulting in sharper and more coherent details (bottom row).

\subsection{Different Backbones and Generative Models}
As shown in \Tref{table:experiment_n3v_other_methods}, we evaluate FillGS with a different generative model, Difix3D+~\cite{wu2025difix3d+}, and a different backbone, Ex4DGS~\cite{lee2024ex4dgs}. FillGS remains effective across these backbones and generators, indicating that its gains are not specific to a particular generative model or backbone.

\begin{table*}[!t]
    \centering
    \begin{adjustbox}{width=1.0\linewidth}
    \small\begin{tabular}{l|rrrrr|rrrrr}
    \renewcommand{\arraystretch}{1.05}
    & \multicolumn{5}{c|}{Interpolation} & \multicolumn{5}{c}{Extrapolation} \\
    & PSNR$\uparrow$ & SSIM$\uparrow$ & LPIPS$\downarrow$ & FID$\downarrow$ & DINOv2$\uparrow$ & PSNR$\uparrow$ & SSIM$\uparrow$ & LPIPS$\downarrow$ & FID$\downarrow$ & DINOv2$\uparrow$ \\\hline
    \multicolumn{1}{l|}{\textit{Generative model: Difix3D+~\cite{wu2025difix3d+}}} & & & & & & \\
    Interpolation-based selection~\cite{kong2025gsgs} & 20.25 & 0.768 & 0.298 & 123.96 & 0.884 & 14.59 & 0.687 & 0.384 & 140.26 & 0.851 \\
    FisherRF-based selection~\cite{Jiang2023FisherRF} & 20.90 & 0.772 & 0.305 & 137.08 & 0.851 & 15.17 & 0.674 & 0.399 & 163.51 & 0.820 \\
    Coverage-based (3D) selection~\cite{kim2025exploregsexplorable3dscene} & 20.88 & 0.774 & 0.298 & 112.15 & 0.884 & 15.46 & 0.676 & 0.397 & 182.84 & 0.803 \\
    Coverage-based (4D) selection~\cite{kim2025exploregsexplorable3dscene} & 21.03 & 0.771 & 0.303 & 125.58 & 0.881 & 15.75 & 0.681 & 0.388 & 151.10 & 0.850 \\
    Ours & \textbf{21.42} & \textbf{0.780} & \textbf{0.280} & \textbf{110.77} & \textbf{0.890} & \textbf{16.03} & \textbf{0.724} & \textbf{0.346} & \textbf{137.17} & \textbf{0.853} \\
    \multicolumn{1}{l|}{\textit{Backbone: Ex4DGS~\cite{lee2024ex4dgs}}} & & & & & & \\
    Ex4DGS & 19.02 & 0.659 & 0.425 & 263.80 & 0.504 & 15.56 & 0.635 & 0.448 & 284.22 & 0.577 \\
    Interpolation-based selection~\cite{kong2025gsgs} & 15.35 & 0.590 & 0.485 & 286.86 & 0.572 & 14.74 & 0.609 & 0.476 & 296.89 & 0.551 \\
    FisherRF-based selection~\cite{Jiang2023FisherRF} & 17.29 & 0.634 & 0.464 & 315.80 & 0.344 & 16.79 & 0.648 & 0.454 & 298.90 & 0.447 \\
    Coverage-based (3D) selection~\cite{kim2025exploregsexplorable3dscene} & 17.68 & 0.641 & 0.452 & 267.81 & 0.542 & 15.85 & 0.640 & 0.445 & 264.16 & 0.573 \\
    Coverage-based (4D) selection~\cite{kim2025exploregsexplorable3dscene} & 18.33 & 0.639 & 0.440 & 270.16 & 0.536 & 15.21 & 0.625 & 0.457 & 259.51 & 0.630 \\
    Ours & \textbf{19.61} & \textbf{0.679} & \textbf{0.378} & \textbf{180.30} & \textbf{0.699} & \textbf{18.07} & \textbf{0.650} & \textbf{0.430} & \textbf{242.47} & \textbf{0.659} \\
    \end{tabular}
    \end{adjustbox}
    \caption{\textbf{Quantitative comparison on Difix3D+~\cite{wu2025difix3d+} (generative model) and Ex4DGS~\cite{lee2024ex4dgs} (backbone) on Neural 3D Video Dataset~\cite{dynerf}.}}
    \label{table:experiment_n3v_other_methods}
\end{table*}

\subsection{Experiments on Monocular Scenes}
\begin{figure*}[t!]
    \centering
    \setlength{\tabcolsep}{0.02cm}
    \setlength{\itemwidth}{2.4cm}
    \renewcommand{\arraystretch}{0.5}
    \begin{adjustbox}{width=1.0\linewidth}
    \hspace*{-\tabcolsep}\small\begin{tabular}{ccccccc}
            \includegraphics[width=1.7777777777\itemwidth]{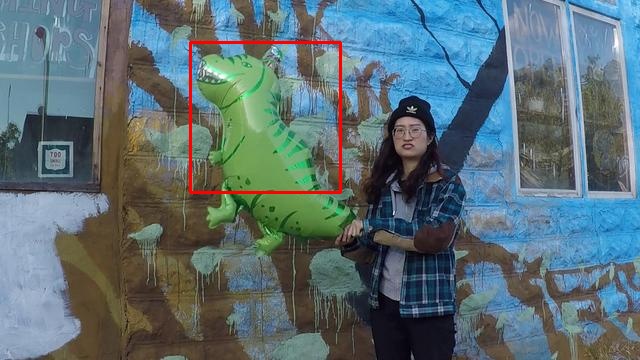} &
            \fboxsep=0pt\fcolorbox{red}{white}{\includegraphics[width=\itemwidth]{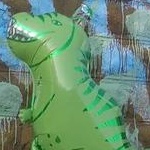}} &
            \includegraphics[width=\itemwidth]{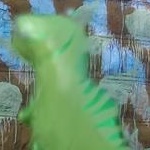} &
            \includegraphics[width=\itemwidth]{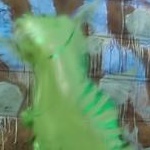} &
            \includegraphics[width=\itemwidth]{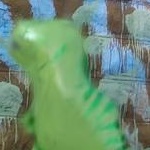} &
            \includegraphics[width=\itemwidth]{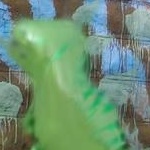} &
            \includegraphics[width=\itemwidth]{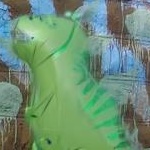} \\
            \includegraphics[width=1.7777777777\itemwidth]{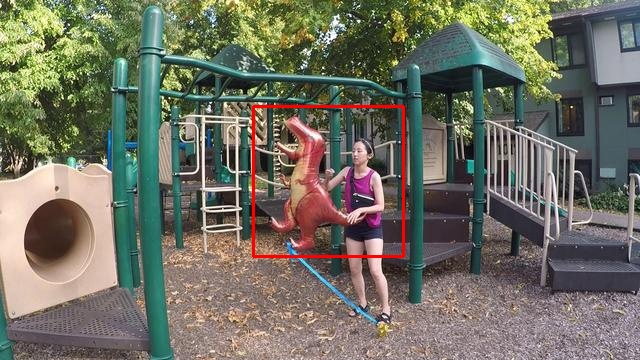} &
            \fboxsep=0pt\fcolorbox{red}{white}{\includegraphics[width=\itemwidth]{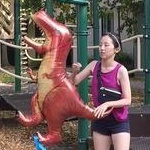}} &
            \includegraphics[width=\itemwidth]{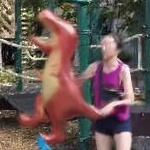} &
            \includegraphics[width=\itemwidth]{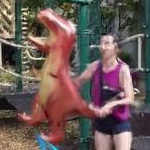} &
            \includegraphics[width=\itemwidth]{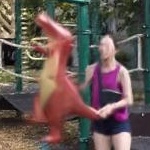} &
            \includegraphics[width=\itemwidth]{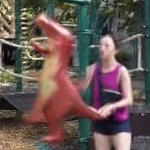} &
            \includegraphics[width=\itemwidth]{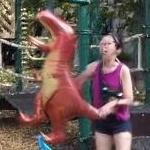} \\
            \includegraphics[width=1.7777777777\itemwidth]{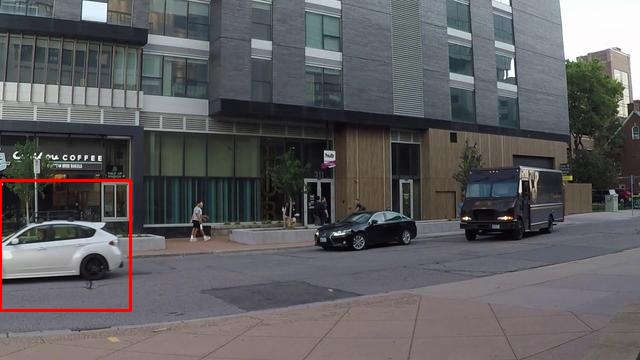} &
            \fboxsep=0pt\fcolorbox{red}{white}{\includegraphics[width=\itemwidth]{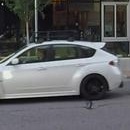}} &
            \includegraphics[width=\itemwidth]{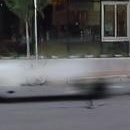} &
            \includegraphics[width=\itemwidth]{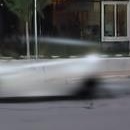} &
            \includegraphics[width=\itemwidth]{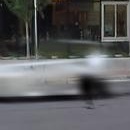} &
            \includegraphics[width=\itemwidth]{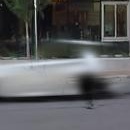} &
            \includegraphics[width=\itemwidth]{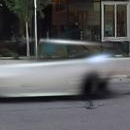} \\
            \multicolumn{2}{c}{Ground-Truth image} &
            \capcell{E-D3DGS~\cite{bae2024ed3dgs}} &
            \capcell{FisherRF-based\\selection~\cite{Jiang2023FisherRF}} &
            \capcell{Coverage-based\\(3D) selection~\cite{kim2025exploregsexplorable3dscene}} &
            \capcell{Coverage-based\\(4D) selection~\cite{kim2025exploregsexplorable3dscene}} &
            \capcell{Ours} \\
        \\
    \end{tabular}\vspace{-1.5em}
    \end{adjustbox}
  \vspace{-0.4cm}\caption{\textbf{Qualitative comparison on the Nvidia Video Dataset~\cite{nvidia}.}}
  \label{fig:exp_nvidia}
\end{figure*}

For further evaluation, we extend seven scenes from the Nvidia dataset~\cite{nvidia} to a monocular setting and conduct evaluations. Specifically, from the 12 cameras, we sample the training camera trajectory by cycling through the eight inner cameras, taking one image per time step. The test cameras are set to the four outer cameras. Quantitative results are reported in \Tref{table:experiment_nvidia}. In monocular settings, where only a single image is available at each time step, our method still outperforms existing view-selection methods. Furthermore, qualitative results are shown in \Fref{fig:exp_nvidia}. Our approach can reconstruct sharper details in dynamic regions.

\begin{table}[!t]
    \centering
    \small\begin{tabular}{l|rrrrr}
    & PSNR$\uparrow$ & SSIM$\uparrow$ & LPIPS$\downarrow$ & FID$\downarrow$ & DINOv2$\uparrow$ \\\hline
    E-D3DGS~\cite{bae2024ed3dgs} & 22.40 & 0.696 & 0.241 & 71.39 & 0.831 \\
    FisherRF-based selection~\cite{Jiang2023FisherRF} & 22.79 & 0.691 & 0.255 & 72.24 & 0.828 \\
    Coverage-based (3D) selection~\cite{kim2025exploregsexplorable3dscene} & 22.70 & 0.704 & 0.227 & 54.40 & 0.857 \\
    Coverage-based (4D) selection~\cite{kim2025exploregsexplorable3dscene} & 22.64 & 0.701 & 0.230 & 55.86 & 0.852 \\
    Ours & \textbf{23.01} & \textbf{0.720} & \textbf{0.216} & \textbf{43.71} & \textbf{0.870} \\
    \end{tabular}
    \caption{\textbf{Quantitative comparison on Nvidia Video Dataset~\cite{nvidia}.}}
    \label{table:experiment_nvidia}
\end{table}

\subsection{Failure Cases}
\begin{figure*}[t!]
    \centering
    \setlength{\tabcolsep}{0.02cm}
    \setlength{\itemwidth}{2.4cm}
    \renewcommand{\arraystretch}{0.5}
    \begin{adjustbox}{width=1.0\linewidth}
    \hspace*{-\tabcolsep}\small\begin{tabular}{cccccccc}
            \includegraphics[width=1.88235294118\itemwidth]{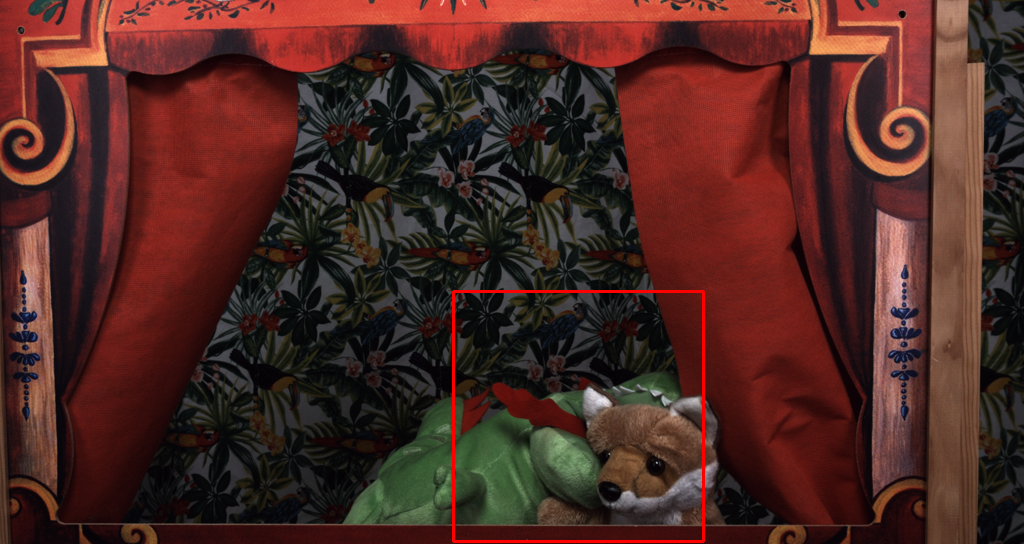} &
            \fboxsep=0pt\fcolorbox{red}{white}{\includegraphics[width=\itemwidth]{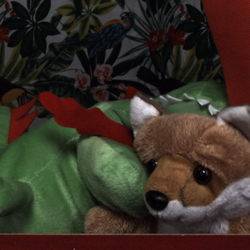}} &
            \includegraphics[width=\itemwidth]{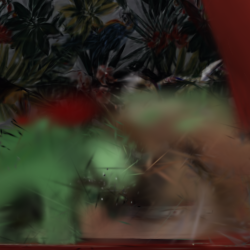} &
            \includegraphics[width=\itemwidth]{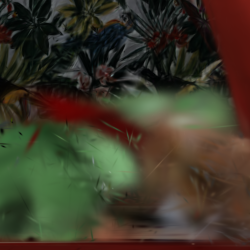} &
            \includegraphics[width=\itemwidth]{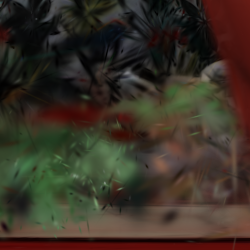} &
            \includegraphics[width=\itemwidth]{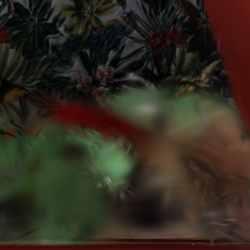} &
            \includegraphics[width=\itemwidth]{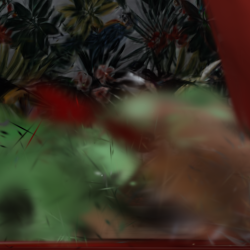} &
            \includegraphics[width=\itemwidth]{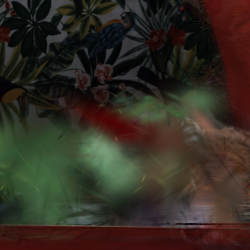} \\
            \multicolumn{2}{c}{Ground-Truth image} &
            \capcell{E-D3DGS~\cite{bae2024ed3dgs}} &
            \capcell{Interpolation-based\\selection~\cite{kong2025gsgs}} &
            \capcell{FisherRF-based\\selection~\cite{Jiang2023FisherRF}} &
            \capcell{Coverage-based\\(3D) selection~\cite{kim2025exploregsexplorable3dscene}} &
            \capcell{Coverage-based\\(4D) selection~\cite{kim2025exploregsexplorable3dscene}} &
            \capcell{Ours} \\
        \\
    \end{tabular}\vspace{-1.5em}
    \end{adjustbox}
  \vspace{-0.4cm}\caption{\textbf{Failure cases.} Our pipeline has difficulty reconstructing complex motions.}
  \label{fig:limitation}
\end{figure*}

Our pipeline may struggle to accurately reconstruct scenes due to the limited representational capacity of the underlying 4DGS representation and the insufficient quality of the generated results. As shown in \Fref{fig:limitation}, this issue is particularly evident in scenes involving complex motion, where accurate reconstruction remains challenging.

\end{document}